%% file: TR/tr_version.tex
\documentclass{article} %

\PassOptionsToPackage{numbers, compress, sort}{natbib}

\usepackage{tcolorbox}
\tcbuselibrary{breakable}

\newtcolorbox{cobeexample}[1]{
  breakable,
  colback=CadetBlue!3,
  colframe=CadetBlue!35,
  colbacktitle=CadetBlue!12,
  coltitle=black,
  fonttitle=\bfseries,
  fontupper=\small,
  title={#1},
  boxrule=0.5pt,
  arc=2pt,
  left=6pt, right=6pt, top=5pt, bottom=5pt,
  before skip=6pt, after skip=8pt
}

\usepackage[utf8]{inputenc} %
\usepackage[T1]{fontenc}    %
\usepackage{hyperref}       %
\usepackage{url}            %
\usepackage{booktabs}       %
\usepackage{amsfonts}       %
\usepackage{nicefrac}       %
\usepackage{microtype}      %
\usepackage{wrapfig}
\usepackage{amsmath}
\usepackage{cleveref}
\usepackage{graphicx}
\usepackage[dvipsnames]{xcolor}
\usepackage{tcolorbox}
\usepackage{enumitem}
\usepackage{tikz}
\usepackage{tikz-3dplot}
\usetikzlibrary{3d,decorations.text}
\usepackage{listings}
\tcbuselibrary{listings,breakable}

\usepackage{booktabs}
\usepackage{tabularx}
\usepackage{makecell}
\usepackage{array}
\usepackage{caption}
\usepackage{adjustbox}
\usepackage{multirow}
\usepackage{longtable}
\usepackage{ragged2e}

\newcolumntype{P}[1]{>{\RaggedRight\arraybackslash}p{#1}}

\usepackage{pifont}

\newcommand{\cmark}{\textcolor{green!60!black}{\ding{51}}} %
\newcommand{\xmark}{\textcolor{red!70!black}{\ding{55}}}   %

\usepackage{inwoo}

\usepackage{titletoc}
\newcommand\DoToC{%
  \setcounter{tocdepth}{3} %
  \startcontents
  {\centering\Large\bfseries Appendix\par}
  \vskip3pt\hrule\vskip5pt
  \printcontents{}{1}{}
  \vskip3pt\hrule\vskip5pt
}

\let\CoBeAddContentsLine\addcontentsline
\usepackage{tr_style,times}
\let\addcontentsline\CoBeAddContentsLine
\iclrfinalcopy

\input{math_commands.tex}

\usepackage{hyperref}
\usepackage{url}
\title{CoBe: A Benchmark For\\Conversational Counterfactual Text Editing}

\author{%
  Arvind Raghavan\thanks{Equal contribution.} \qquad Inwoo Hwang$^\ast$ \qquad Shreyas Havaldar$^\ast$ \\
  \textbf{Kai-Zhan Lee} \qquad 
  \textbf{Aurghya Maiti} \qquad \textbf{Yushu Pan} 
  \\
  \textbf{Jeffrey Wu}  \qquad \textbf{Mingxuan Li} \qquad \textbf{Elias Bareinboim} \vspace{+0.06in}\\
  Causal Artificial Intelligence Lab \\
  Columbia University \\
  \texttt{\{ar, inwoo, shreyas, kl, aurghya, yushupan, ml, eb\}@cs.columbia.edu}\\
  \texttt{jgw2140@columbia.edu}
}

\begin{document}

\begin{NoHyper}
\maketitle
\end{NoHyper}

\begin{abstract}
\textit{Counterfactual text editing} is the ability to revise a given text in light of a hypothetical change while preserving facts not causally affected by that change. When asked what would have happened if a doctor had not recommended surgery, the model output should retain the patient's medical history and modify only the consequences of the decision. This requires causal understanding: a model must determine, without being explicitly instructed to, what events are affected by the hypothetical change and what should remain fixed. This skill is central to human reasoning, especially in safety-critical domains like healthcare or law. By contrast, the task of \textit{associational text generation} merely generates text statistically associated with the altered condition. In the medical example, an associational response might make the patient's symptoms less severe because milder cases correlate with no surgery. Such responses produce a plausible scenario, but not the counterfactual version of the given one. Prior counterfactual-text datasets do not clearly separate these modes of reasoning. We introduce the \textbf{{Co}}unterfactual \textbf{Be}nchmark (\texttt{CoBe}), a testbed for models' zero-shot ability to counterfactually edit a given textual scenario when prompted to do so in a \textit{conversationally natural} way. \texttt{CoBe} contains 2000 questions spanning healthcare, science and engineering, business, and everyday scenarios. We test a suite of frontier and smaller models, with frontier systems achieving an average score of {58.05\%}. Our findings indicate that models frequently default to associational generation rather than preserving the causal structure of the original scenario. Moreover, performance does not scale consistently with either model size or output token length, suggesting that the failure may reflect model training objectives rather than insufficient model capacity.
\end{abstract}

\input{TR/tr_introduction}

\vspace{-0.05in}
\input{relatedwork}

\input{TR/tr_methods}
\input{TR/tr_eval}

\input{TR/tr_results}

\input{TR/tr_conclusion}

\subsection*{AI use statement}

In this work, we used generative AI tools as part of the dataset generation and evaluation pipeline, as detailed in Section 3 and 4 in the paper. Specifically, LLMs were used to augment an author-curated set of core scenarios by generating variations to increase diversity and LLMs were also used as a part of the benchmark evaluation pipeline to assess whether model-generated counterfactual edits satisfied the evaluation criteria. All dataset examples went through several stages of human review and the reliability of the LLM-evaluation pipeline also underwent human agreement experiments to ensure correctness. We additionally used LLMs for assistance in code writing and debugging, and for paper editing tasks like improving phrasing. All AI-assisted material was reviewed by the authors, and the codebase and experimental results were independently and thoroughly checked by the authors before inclusion. We did not rely on generative AI to draw the insights and conclusions highlighted in our work. We take full responsibility for the final content of this work, including text, claims or artifacts produced with the aid of generative AI.

\bibliography{references.bib}
\bibliographystyle{iclr2027_conference}

\clearpage
\appendix
\DoToC
\input{TR/tr_appendix}
\stopcontents

\end{document}

%% file: math_commands.tex
\usepackage{amsmath,amsfonts,bm}

\def\eqref#1{equation~\ref{#1}}

\def\1{\bm{1}}

\DeclareMathAlphabet{\mathsfit}{\encodingdefault}{\sfdefault}{m}{sl}
\SetMathAlphabet{\mathsfit}{bold}{\encodingdefault}{\sfdefault}{bx}{n}

%% file: TR/tr_introduction.tex
\begin{wrapfigure}{r}{0.65\linewidth}
    \centering
    \vspace{-0.55in}
    \includegraphics[width=\linewidth]{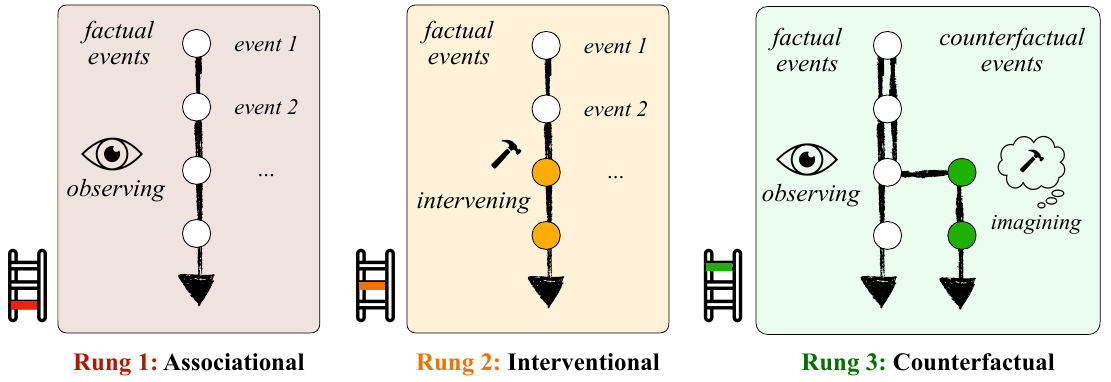}
    \vspace{-0.15in}
    \caption{The three rungs of Pearl's causal ladder are (i) associational; (ii) interventional; and (iii) counterfactual.}
    \vspace{-0.1in}
    \label{fig:pch}
\end{wrapfigure}

\section{Introduction} \label{sec: intro}

Conversational counterfactuals are a vital tool for understanding the world and the consequences of choosing different actions in a given scenario. Consider a classic example in the philosophy of language: the statements (1) \textit{If Oswald did not shoot Kennedy, someone else did}; and (2) \textit{Had Oswald not shot Kennedy, someone else would have}. Most readers would agree with the first statement while disagreeing with the second, intuitively grasping that the two make fundamentally different claims. The first is a factual claim, whereas the second is a \textit{counterfactual} claim about what \textit{would have happened} had we intervened in the given scenario with a different condition. 

Counterfactual reasoning belongs to the third rung of Judea Pearl's celebrated \textit{ladder of causation}, a.k.a the {Pearl Causal Hierarchy}, or PCH \citep{pearl:mackenzie2018}\citep[Ch.~5]{bareinboim:textbook}. Rung 1 involves observing statistical \textit{associations} among events in the factual world. Rung 2 involves \textit{intervening} in the factual world and observing outcomes. Rung 3 involves observing factual events and reasoning about what would \textit{counterfactually} have happened had we intervened in the given scenario. Notably, a counterfactual only alters events causally downstream of the hypothetical intervention, and keeps fixed background factors in the original scenario that are causally prior to the intervention (Fig. \ref{fig:pch}). Refer to App. \ref{sec:pch} for a primer on causal models. Counterfactual reasoning is well-established to be central to personalized decision-making \citep{bareinboim:etal15, MuellerPearl_2023, raghavan2025realizability}, mediation analysis \citep{pearl:01, rubin:04, avin:etal05}, fairness analysis \citep{zhang:bar18a, plecko:bareinboim24}, and explainability \citep{halpern:pea99,leekz:nte2025}.

In this work we focus on \textit{conversational counterfactuals}: Rung 3 questions articulated in everyday colloquial language, using the subjunctive "would have...had" grammatical construction. Would the patient have succumbed to an infection had she not been moved to the crowded ward? Would oil prices be lower today if the UAE had quit OPEC earlier? Conversational counterfactuals are integral to how humans think, communicate, and construct explanations for why events occurred; they are foundational to our intuitions about ethics, regret, credit, and responsibility \citep{pearl:mackenzie2018,pearl:2k,lewis:73a,hume:1748,bareinboim:textbook,gopnik:etal04}. We next formally define this important capability that we wish to test on language models.

\begin{figure}[t]
    \centering
    \includegraphics[width=0.95\linewidth]{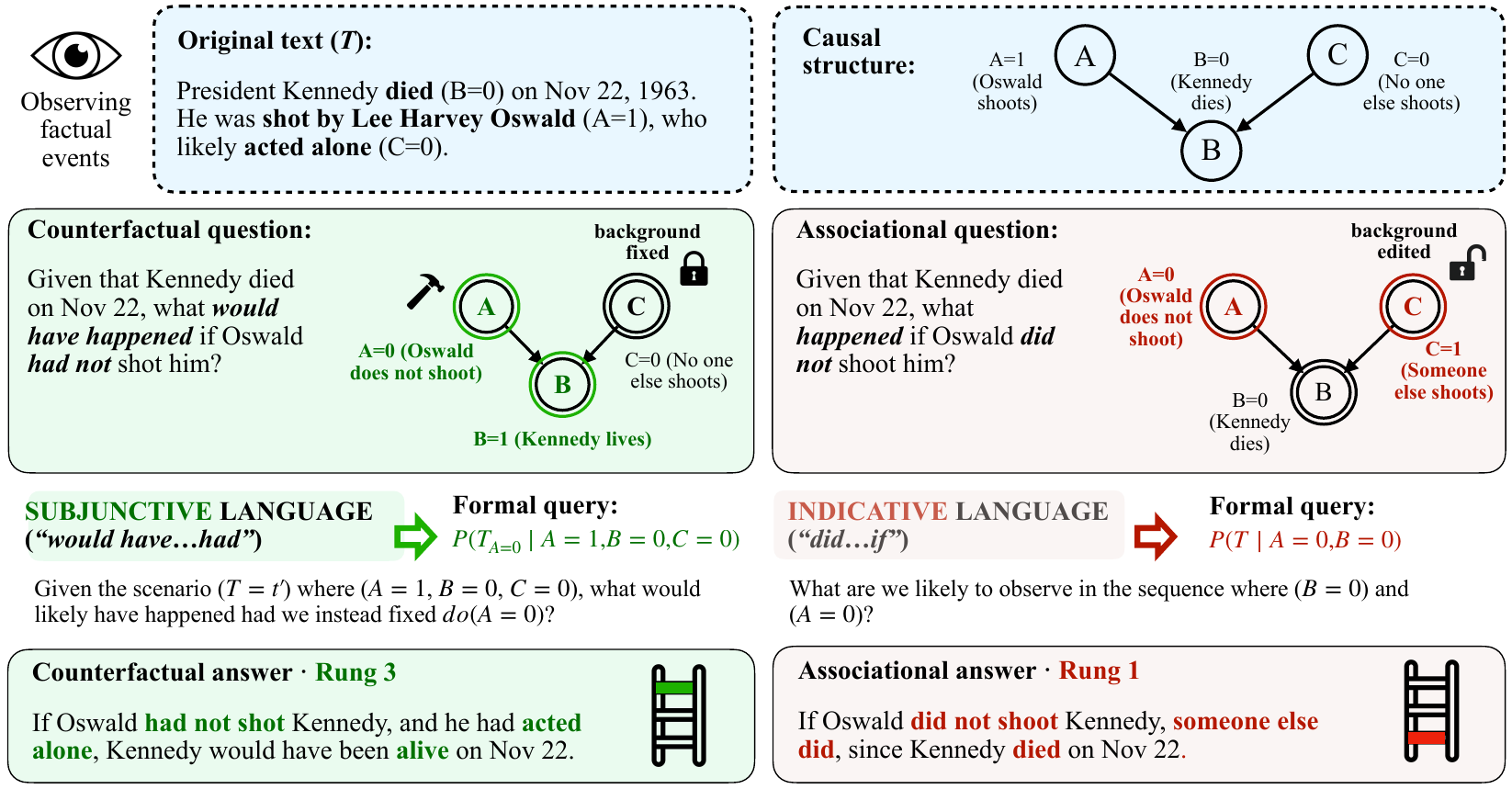}
    \vspace{-0.in}
    \caption{Counterfactual text editing, belonging to Rung 3 of Pearl's \textit{causal ladder}, \textcolor{OliveGreen}{keeps background factors constant}, and reasons about \textcolor{OliveGreen}{what would have happened as the result of a hypothetical intervention}. By contrast, associational text generation belongs to {Rung 1}, and generates \textcolor{BrickRed}{alternative scenarios correlated with the altered condition (possibly changing background factors)}. \textbf{Note}: counterfactual scenarios need not be realistic or likely to occur, ordinarily.}
    \vspace{-0.2in}
    \label{fig:ctf_vs_association}
\end{figure}
\vspace{-0.1in}

\vspace{-0.05in}
\paragraph{Conversational counterfactual text editing.} A language model receives two inputs: (i) a textual story $T$ describing a set of events $(A, B...)$ where some events causally affect other events in the story; and (ii) a conversational counterfactual question involving a hypothetical intervention in this particular storyline, phrased in colloquial subjunctive "would have...had" grammar. From this prompt, the model is expected to infer and execute three steps: \textit{(a) keep fixed any background factors/events in the story that are not causally affected by the intervened-variable in the original timeline; (b) enact the requested intervention by imagining an externally enforced edit to the storyline; and (c) reason about the commonsense downstream effects on subsequent events in the storyline due to enforcing this change, and produce edited text reflecting this counterfactual storyline.} 

Example shown in Fig. \ref{fig:ctf_vs_association} in green. We also provide a detailed illustration, and additional formal background on conversational counterfactuals and the subjunctive conditional construction in App. \ref{sec:subjunctives}, using this Oswald/Kennedy example. Note: in our definition of conversational counterfactuals, we intentionally avoid any technical instructions about causal graphs, counterfactuals, or step-by-step procedures in our prompts. 
Our motivation for focusing on {conversational} prompting is twofold. First, this is a natural human capability: when asked a question like \textit{"what would have happened in this scenario if I had done something different?"} we automatically understand that this means the intervention should only affect downstream events holding fixed all background factors and prior events, without requiring causal formalisms or detailed instructions such as "keep prior events fixed". Second, AI assistants are increasingly embedded in everyday and professional workflows \citep{anthropic2026cowork}, with personal agents such as Meta's Muse explicitly designed for communication as users would with another person \citep{meta2026muse}. In these settings, users should be able to express a hypothetical change in ordinary interactive language without needing to detail in each case which variables should remain invariant, or requiring expertise in task-specific prompt tuning \citep{zamfirescu2023johnny}. Our task is therefore designed to test whether models can \textit{infer} these edit requirements from the scenario and conversational request, without requiring the user to supply a detailed causal task recipe. App. \ref{sec: alt_prompts} examines more explicit prompt instructions to distinguish performance in the conversational setting from performance with additional procedural guidance. 

\begin{figure}[!t]
    \centering
    \includegraphics[width=\linewidth]{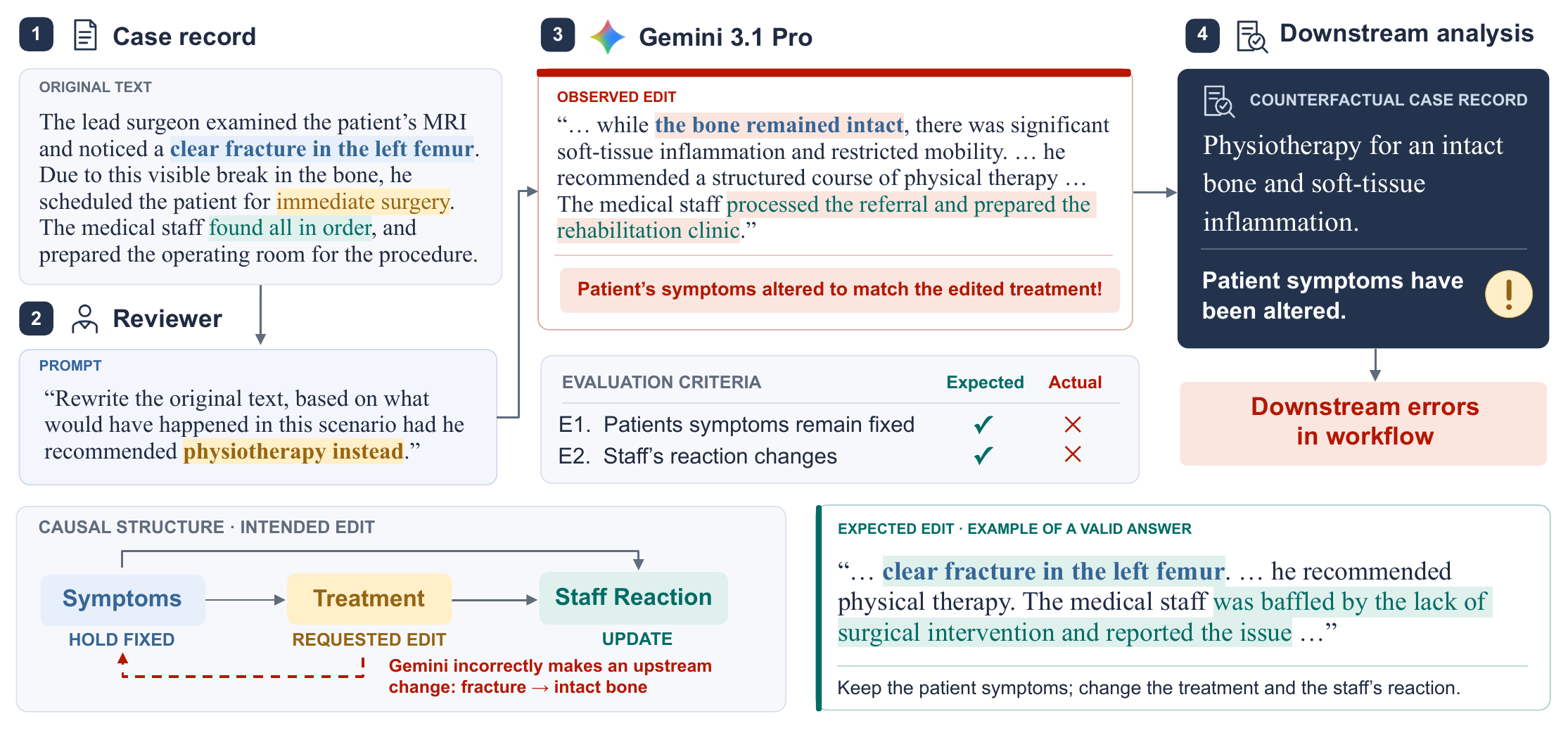}

    \vspace{-7pt}
    \caption{Idealized healthcare workflow depicting a staff reviewing medical records to generate case studies for training. Gemini 3.1 is prompted, in conversational language, to incorporate a counterfactual edit to patient treatment. Gemini incorrectly alters patient symptoms, causing downstream errors.}
    \label{fig:failures}
    \vspace{-0.2in}
\end{figure}

Unfortunately, a growing body of work documents that LLMs often rely on \textit{associational} signals rather than genuine causal structure, learning surface heuristics from their pretraining distribution \citep{wu2024causalitylargelanguagemodels,joshi2024llmspronefallaciescausal} and often learning shortcuts to predict ``right but for the wrong reasons'' \citep{geirhosshortcutlearning_2020,du2023shortcut,mccoy2019rightforwrong}. In contrast to counterfactual edits, the task of \textbf{associational generation} (belonging to Rung 1 of the causal ladder) \textit{generates text by merely conditioning on an alternative event to ask what scenarios are statistically likely to co-occur with the given change}. We illustrate this in Fig. \ref{fig:ctf_vs_association} in red, and expand in App. \ref{sec:subjunctives}. 

Given how embedded language models are in modern workflows, confusing counterfactual and association generation can lead to errors in safety-critical domains like medicine, law etc. We show in Fig. \ref{fig:failures} an idealized healthcare workflow where a user is interacting with an AI assistant to generate case studies for the same patient under different treatments. Frontier models incorrectly alter the patient history in the output, introducing downstream errors in the workflow. Such shortcut learning based on spurious correlations in the training data risks systematic, hard-to-predict errors \citep{kiciman_newfrontier_2024,Kim2025medical}. For instance, when trying to mitigate bias using counterfactual data augmentation, naive approaches can introduce new biases due to spurious correlations in the data \citep{mouli2022biaschallenges,tokpo_fairflow24,jin2025coba}.

With this motivation, we introduce \textbf{Co}unterfactual \textbf{Be}nchmark (\texttt{CoBe}): a dataset for testing the {native} ability of language models to produce causally-valid counterfactual edits to text, when prompted to do so in \textit{conversational} language, without specialized prompts. The key highlights are as follows:
\begin{itemize}[leftmargin=15pt]
\vspace{-0.075in}
    \item We curate a set of \textbf{$2000$ samples} ($200$ public), each containing, in natural language, (i) a short text describing a scenario, and (ii) a set of single-turn prompts requesting, in colloquial everyday language, a \textbf{counterfactual edit} to the paragraph.
     \vspace{-0.03in}
    \item We provide the \textbf{evaluation criteria} for each question
    to determine the validity of model responses, offering a principled and causally-grounded way of evaluating counterfactual edits,
    verified by human experts.
     \vspace{-0.03in}
    \item Our dataset covers scenarios from \textbf{diverse domains}: Science \& Research, Engineering, Business \& Services, Everyday Scenarios, Healthcare, and Legal, Policy \& HR, among others.
     \vspace{-0.03in}
    \item We investigate a comprehensive suite of closed and open-source models, with results summarized in Sec.~\ref{sec: eval}. Notably, \textbf{frontier models only score an avg. of 58.05\%} as of July 2026.
    \vspace{-0.03in}
    \item We taxonomize the failure modes we observed and characterize the idiosyncratic error profiles of different models. For instance, we observe that Gemini-3.1-Pro outputs appear to strongly favor correlational consistency, and GPT-5.4-Pro struggles to incorporate background information from the scenario in its edits. 
\end{itemize}
\vspace{-0.075in}
To our knowledge, we provide the first testbed for zero-shot conversational counterfactual edits, grounded in formal causal semantics, in the domain of narrative stories. Our evaluation criteria allow for multiple valid responses that respect counterfactual reasoning. We are also the first to characterize the distinct error-profiles of different models on this important task. All of these allow \texttt{CoBe} to be seamlessly integrated as a training signal into pipelines designed to improve model reasoning, both fine-tuning and inference-time. 

Counterfactual text editing is a specific manifestation of a broader causal reasoning capability. Causal intelligence spans many activities, including causal understanding, explanation, decision-making, generalization, representation learning, and causal simulation \citep[Ch.~1]{bareinboim:textbook}. \texttt{CoBe} focuses on a very specific sub-task: we test whether models can infer which facts to preserve or revise from the narrative text and a conversational edit request, without explicit instructions specifying causal dependencies, invariant facts, or a counterfactual reasoning procedure.

%% file: relatedwork.tex
\section{Related Work}
\vspace{-0.05in}

\begin{wraptable}{r}{0.63\textwidth}
\vspace{-0.475in}
\centering
\caption{Comparison to prior benchmarks.}
\label{tab:related-work-wrapped}
\scriptsize
\setlength{\tabcolsep}{2pt}
\renewcommand{\arraystretch}{1.12}

\begin{tabular}{@{}p{0.24\linewidth}ccccc@{}}
\toprule
\textbf{Prior work}
& \makecell{\textbf{Ctf-editing}\\\textbf{3 steps}}
& \makecell{\textbf{Narrative}\\\textbf{scenarios}}
& \makecell{\textbf{No causal}\\\textbf{structure}}
& \makecell{\textbf{Open-ended}\\\textbf{generation}}
& \makecell{\textbf{Expert/formal}\\\textbf{verification}} \\
\midrule
TimeTravel  & \cmark & \cmark & \xmark & \cmark & \xmark \\
CRASS 
& \xmark & \cmark & \cmark & \xmark & \xmark \\
CLadder  & \cmark & \xmark & \xmark & \xmark & \cmark \\
CounterBench  & \cmark & \xmark & \xmark & \xmark & \cmark \\
\midrule
\textbf{\texttt{CoBe} (our work)} & \cmark & \cmark & \cmark & \cmark & \cmark \\
\bottomrule
\end{tabular}

\vspace{-1em}
\end{wraptable}

We group related work under common themes for discussion. In Table \ref{tab:related-work-wrapped}, we compare notable datasets to \texttt{CoBe} across five dimensions: does the benchmark (1) ground itself in the formal 3-step definition of counterfactual text editing (see Sec. \ref{sec: intro}); (2) test understanding of narrative stories (vs. mathematics or coding questions phrased in text); (3) withhold information about the story's causal structure (i.e., not providing a causal graph, nor pre-specifying which sentences need to remain invariant); (4) evaluate open-ended generation (vs. multiple choice selection); and (5) include evaluation criteria verified by experts or formal solvers (vs. crowd-sourced).

\textbf{General benchmarks.} It is now standard for model cards to describe performance on general benchmarks like MMLU \cite{Hendrycksetal2020}, BigBench \cite{Srivastavaetal2022}, GSM8K \cite{cobbe2021trainingverifierssolvemath}, ARC-AGI-2 \cite{chollet2025arcagi2} etc., which test a combination of domain knowledge, and abstract and common-sense reasoning skills. Some of these baselines do occasionally test counterfactual reasoning skills, albeit in an indirect and non-rigorous way, pointing to a gap in the literature of a test for making formally-grounded and open-ended counterfactual edits to a text.

\textbf{Counterfactual story rewriting.} \textit{TimeTravel} \cite{qin2019counterfactual} and \textit{PASTA} \citep{ghosh_pasta} test the ability to rewrite a simple story based on a counterfactual edit. Both are built on the dated \textit{ROCstories} dataset \citep{mostafazadeh-rocstories}: a restrictive corpus of short everyday stories that follow a simple linear timeline, often with a single event/actor, where story structure makes it clear that only sentences following an edit should be changed. Both benchmarks use crowd-sourced judgments and similarity metrics like BLEU for evaluation, which later work notes only correspond weakly with counterfactual validity \citep{li_click23}. \texttt{CoBe}'s evaluation criteria are expert-verified, allowing multiple valid rewrites without requiring similarity to a reference answer. \citet{mu-li-2024-causal} propose a method for counterfactual story rewrites, but they also assume the task pre-specifies which parts of the story need to remain invariant.

\textbf{Other causal reasoning tasks.}
\textit{CLadder} \citep{cladderjin23} and
\textit{CounterBench} \citep{chen2025counterbench} evaluate
causal queries over explicitly specified causal structures
expressed in natural language.
\textit{Corr2Cause} \citep{jin2024corr2cause} tests whether
causal hypotheses follow from supplied statistical dependence
and independence relations.
Other benchmarks assess causal understanding in text:
\textit{CRASS} \citep{frohberg-binder-2022-crass} tests
multiple-choice questions about natural-language counterfactuals;
\textit{CRAB} \citep{romanou-etal-2023-crab} assesses the
strength of causal relationships between events in news
narratives; 
\textit{CausalT5k} \citep{geng2026causalt5k} diagnoses causal
reasoning failures in natural-language scenarios, including
counterfactual validity and appropriate refusal under
insufficient evidence.
These tasks complement \texttt{CoBe}, which evaluates
narrative revision from conversational counterfactual requests,
requiring models to infer which facts to preserve and which
consequences to update without a supplied causal graph or
explicit causal editing instructions.

%% file: TR/tr_methods.tex
\vspace{-0.05in}
\section{\texttt{CoBe}: A Dataset for Counterfactual Text Editing} \label{sec: data}
\vspace{-0.05in}

In this section, we discuss dataset construction, the dataset schema, key statistics, and our evaluation methodology for model responses to each sample.

\input{dataset_statistics}

\vspace{-0.1in}
\paragraph{Dataset construction.}
\label{subsec: data construction}
We constructed \texttt{CoBe} in four stages
(App.~\ref{appendix:QA}). First, student researchers who had
completed a graduate causal inference course contributed
425 candidate examples. Second, the authors reviewed and
refined promising submissions into 106 core examples spanning
different domains, causal structures, and potential reasoning
failures. Third, LLM-assisted augmentation produced a pool of
25 variations per core example, with the goal of adding domain and linguistic variety
while preserving the intended causal
relationships. Fourth, author review and external expert
feedback informed the final filtering, removing 650 of the
2,650 augmented examples and yielding 2,000 retained samples.

\vspace{-0.1in}
\paragraph{Filtering and quality assurance.}
At stages two and four, each example
was reviewed by at least two authors for story coherence,
clarity of the intervention, and the soundness of its
preservation and update criteria. Each example required both authors to support inclusion in the dataset, with disagreements resolved through discussion.
Following initial author filtering, three causal inference
experts uninvolved in construction independently assessed
a set of 100 random samples for coherence,
grammar, and validity of the counterfactual
edit. Their feedback identified a systematic inconsistency
and prompted further exclusions. App.~\ref{appendix:QA}
details the review procedures and audit results.

\vspace{-0.1in}
\paragraph{Dataset schema.}
\label{subsec: data schema} Each sample in the dataset contains the following fields. We provide an instantiated sample in App. \ref{subsubsec: sample json}.

\vspace{-0.05in}
\begin{itemize}[leftmargin=0.5cm]
\vspace{-0.05in}
    \item \texttt{Story}: a natural-language text containing a story describing a series of events.
    \item \texttt{Domain}: a metadata tag used to group scenarios into domains, summarized in Table \ref{tab:dataset-stats-domain}.
    \item \texttt{Query}: a counterfactual edit request, in conversational language, to change some detail in the story and produce the counterfactual text; described in more detail below.
    \item \texttt{Illustrative answer (not scored)}: an author-crafted example of a valid response to the query, for reference (multiple valid answers are possible); this is not scored.
    \item \texttt{Evaluation criteria}: the criteria used to judge whether a model response is a valid counterfactual edit to the text, described in more detail in Sec. \ref{subsubsec: et}.
\end{itemize}
\vspace{-0.1in}

\begin{figure}[t!]
    \centering
    \includegraphics[width=\linewidth]{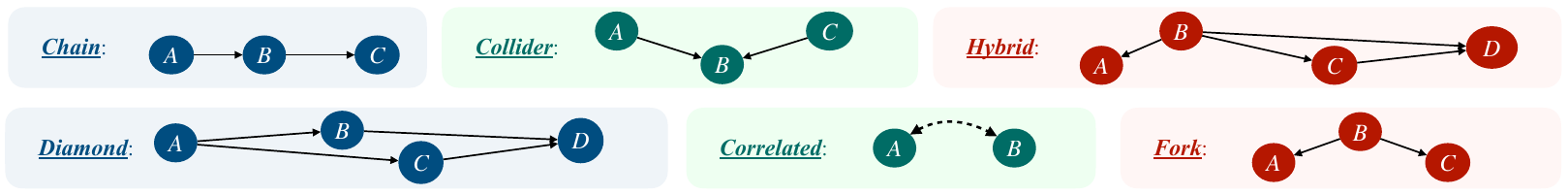}
    \caption{The types of underlying graphs that determine the interaction of variables in a story.}
    \vspace{-0.2in}
    \label{fig:graph_types}
\end{figure}

\vspace{-0.05in}
\paragraph{Query.} Following the task definition in Sec. \ref{sec: intro}, a counterfactual edit query asks what would the edited text read if we were to hypothetically intervene in the original story described in the text to enforce a change, and allow the change to affect subsequent events in the story. We phrase this request in three ways per scenario, using the following syntax:
\vspace{-0.05in}
\begin{itemize}[leftmargin=10pt]
    \item [-] (A) \textit{``Rewrite the original text, based on what would have happened
instead had...<intervention>.''}
    \item [-] (B) \textit{``Based on the preceding text, rewrite the scenario to reflect
what would have happened if ...<intervention had occurred>.''}
    \item [-] (C) \textit{``Rewrite the above passage to illustrate what would have occurred
had ...<intervention>.''}
\end{itemize}
\vspace{-0.05in}
These phrasings were crafted to elicit the interpretation of a counterfactual edit, following standard subjunctive {``would have''} grammar. We provide detailed formal background on how such counterfactual statements are to be interpreted, and the role of subjunctive conditions in App. \ref{sec:subjunctives}. By averaging scores for responses to all three queries per scenario, we reduce any prompt-related ambiguity that the question is intended to be interpreted counterfactually, in the sense of the 3-step definition in Sec.~\ref{sec: intro}. We also verify human agreement with our prompt interpretation and evaluation criteria in App. \ref{appendix:human}.

We discussed extensively in Sec. \ref{sec: intro} our motivation for keeping these prompts conversational, and avoiding any causal terminology or explicit task instructions. We provide prompt ablations studies in App. \ref{sec: alt_prompts}, where we test models using increasingly more detailed prompts that provide procedural guidance for this task, while making a clear scope separation from our task setting.

\vspace{-0.1in}
\paragraph{Graph variety.} Unlike previous counterfactual text editing datasets, \texttt{CoBe} ensures variety across different causal structures connecting events in each story, as listed in Fig. \ref{fig:graph_types} and \Cref{tab:dataset-stats-graph}.

\vspace{-0.1in}
\paragraph{Samples.}
We provide reference samples from our dataset in App. \ref{sec:examples}. Each sample in the dataset is numbered as \texttt{XXXvYY}, where \texttt{XXX} corresponds to the original {core set} example ID that was used as an inspiration to generate this variation (\texttt{v}), and \texttt{YY} represents the variation ID for that particular sample among all the other variations generated for the \texttt{XXX}th {core set} ID, indexed starting from $1$, for example \texttt{307v8}. We also provide access to sample data representing 10\% of our dataset, in this linked \href{https://kaggle.com/datasets/5ae0d5bcf4d8f3ed1af4aae45466f2b49d4b21a3d40783980a2bd3b315437204}{Kaggle repository}.

\vspace{-0.05in}
\subsection{Evaluation Methodology} \label{subsubsec: et}
\vspace{-0.05in}

For each story in the dataset, the \texttt{Evaluation criteria} are crafted to check whether a model's response to the query is a valid counterfactual edit under the benchmark's intended commonsense interpretation. These criteria are grounded in the three steps of the counterfactual text editing task formally defined in Sec.~\ref{sec: intro}.

The first two strings in \texttt{Evaluation criteria} specify story-specific constraints. Together with a separate system-level check of connectors, these identify three error types in the model response. Each error represents a violation of the intended counterfactual edit. Examples of each error type are linked in parentheses:

\begin{enumerate}[leftmargin=15pt]
\vspace{-0.1in}
    \item \textbf{Wrongly editing unaffected facts}: The first string in \texttt{Evaluation criteria} lists facts in the original scenario that should remain unchanged in the model response. These include events that are neither directly intervened on nor causally downstream of the intervention, as well as facts that remain unchanged despite the intervention in the given scenario (e.g., \hyperref[subsubsec: fm1]{403v1}).

    \vspace{-0.033in}
    \item \textbf{Failing to make required changes}: The second string in \texttt{Evaluation criteria} lists original facts that should change to reflect the requested intervention and its consequences. These changes are specified for the particular scenario; being causally downstream does not, by itself, require a variable's value to change (e.g., \hyperref[subsubsec: fm3]{309v26}).

    \vspace{-0.033in}
    \item \textbf{Wrong connectors}: A separate system-level check assesses whether connector words such as \textit{because of, since, so, but} appropriately reflect the relationships in the rewritten scenario. For instance, retaining \textit{due to} can incorrectly present an unchanged danger as the reason for a newly stipulated decision, where \textit{despite} would express the intended contrast (e.g., \hyperref[subsubsec: fm6]{115v3}). This check doesn't require any particular replacement words, but rather concerns the relationship expressed.
\end{enumerate}

\vspace{-0.1in}
The \texttt{Illustrative answer (not scored)} provides one valid rewrite but is not used for reference matching. Multiple responses may satisfy the evaluation criteria. We discuss more fine-grained failure modes in Sec.~\ref{subsubsec: fm}.

\paragraph{Pipeline.}
We first generate responses for each combination of \texttt{Story}
and \texttt{Query}, as depicted in \Cref{fig:pipeline}. We then use
Claude-Opus-4.7 as an LLM judge to evaluate each response using three
checks. The first two assess the supplied \texttt{Evaluation criteria}:
facts that should remain true in the rewrite (E1), and original facts
that should change (E2). A separate check assesses causal connector
consistency (E0). Each check returns \texttt{T} if its requirements
are satisfied and \texttt{F} otherwise; a response is marked correct
only if all three checks pass. Thus, \texttt{T} denotes satisfaction
of a check, rather than the truth of every original statement.

\paragraph{Evaluation rationale.}
Counterfactual editing admits multiple valid rewrites, making
agreement with a reference answer an incomplete measure of
correctness. Reference-based metrics such as \textit{BLEU}
\citep{10.3115/1073083.1073135}, \textit{ROUGE}
\citep{lin-2004-rouge}, and \textit{MoverScore}
\citep{zhao-etal-2019-moverscore} do not directly test which facts
should remain invariant or change under the intervention (a major limitation in prior benchmarks).
Our evaluation instead checks these scenario-specific constraints
and the consistency of causal language, allowing different
wordings and outcomes that satisfy the criteria.

\paragraph{Evaluation limitation.}
The implementation of E2 accepts the removal of a listed fact as
well as its modification. Consequently, an incomplete rewrite can
pass this check by omitting a consequence that should have been
updated. We did not observe systematic use of this shortcut in the
responses we manually inspected. Moreover, the expert evaluation
of 250 sampled responses in App.~\ref{appendix:human} provides
evidence of agreement between the judge and human ground truth assessments.
This supports the final results as indeed representative of the model's actual
performance on this task. We do not claim our evaluation criteria are robust to deliberate omission or other
strategies designed to satisfy the checks without completing
the editing task.

\paragraph{Human agreement.} We conduct human agreement experiments to assess the reliability of our evaluation pipeline, with three independent expert human raters.
Our LLM-Judge shows a high agreement (Cohen's $\kappa$ = 0.812) with human raters, demonstrating the soundness of our evaluation criteria, and sufficient statistical confidence to scale the evaluator across the broader benchmark.
See App.~\ref{appendix:human} for details. 

%% file: dataset_statistics.tex
\begin{figure}[t]
\centering

\begin{minipage}[t]{0.32\linewidth}
\centering
\captionof{table}{Dataset statistics.}
\vspace{-7pt}
\label{tab:dataset-stats}
\small
\setlength{\tabcolsep}{4pt}
\renewcommand{\arraystretch}{1.15}
\begin{adjustbox}{width=0.9\textwidth}
\begin{tabular}{@{}lc@{}}
\toprule
\textbf{Statistic} & \textbf{Value} \\
\midrule
No. of scenarios & 2000 \\
No. of domains & 7 \\
Avg. words/scenario & 39 \\
Avg. sentences/scenario & 2.35 \\
Avg. clauses/scenario & 5.43 \\
Avg. clauses/sentence & 2.44\\

\bottomrule
\end{tabular}
\end{adjustbox}
\end{minipage}%
\hfill
\begin{minipage}[t]{0.35\linewidth}
\centering
\captionof{table}{Domain types.}
\vspace{-7pt}
\label{tab:dataset-stats-domain}
\small
\setlength{\tabcolsep}{4pt}
\renewcommand{\arraystretch}{1.15}
\begin{adjustbox}{width=.62\textwidth}
\begin{tabular}{@{}lc@{}}
    \toprule
    \textbf{Career Domain} & \textbf{\%} \\
    \midrule
    Science \& Research & 16.0 \\
    Engineering & 15.7 \\
    Business \& Services & 14.5 \\
    Everyday & 14.2 \\
    Healthcare & 6.0 \\
    Legal, Policy \& HR & 4.7 \\
    Others & 28.9 \\
    \bottomrule
  \end{tabular}
\end{adjustbox}
\end{minipage}%
\hfill
\begin{minipage}[t]{0.28\linewidth}
\centering
\captionof{table}{Graph types.}
\vspace{-7pt}
\label{tab:dataset-stats-graph}
\small
\setlength{\tabcolsep}{4pt}
\renewcommand{\arraystretch}{1.15}
\begin{adjustbox}{width=0.7\textwidth}
\begin{tabular}{@{}lc@{}}
\toprule
\textbf{Graph} & \textbf{\%} \\
\midrule
Chain-like & 51.0\% \\
Collider-like & 21.9\% \\
Correlated & 5.7\% \\
Diamond-like & 4.5\% \\
Fork-like & 4.8\% \\
Hybrid & 11.9\% \\
\bottomrule
\end{tabular}
\end{adjustbox}
\end{minipage}
\vspace{-0.37cm}
\end{figure}

%% file: TR/tr_eval.tex
\begin{figure}[t!]
    \centering
    \includegraphics[clip, width=\linewidth]{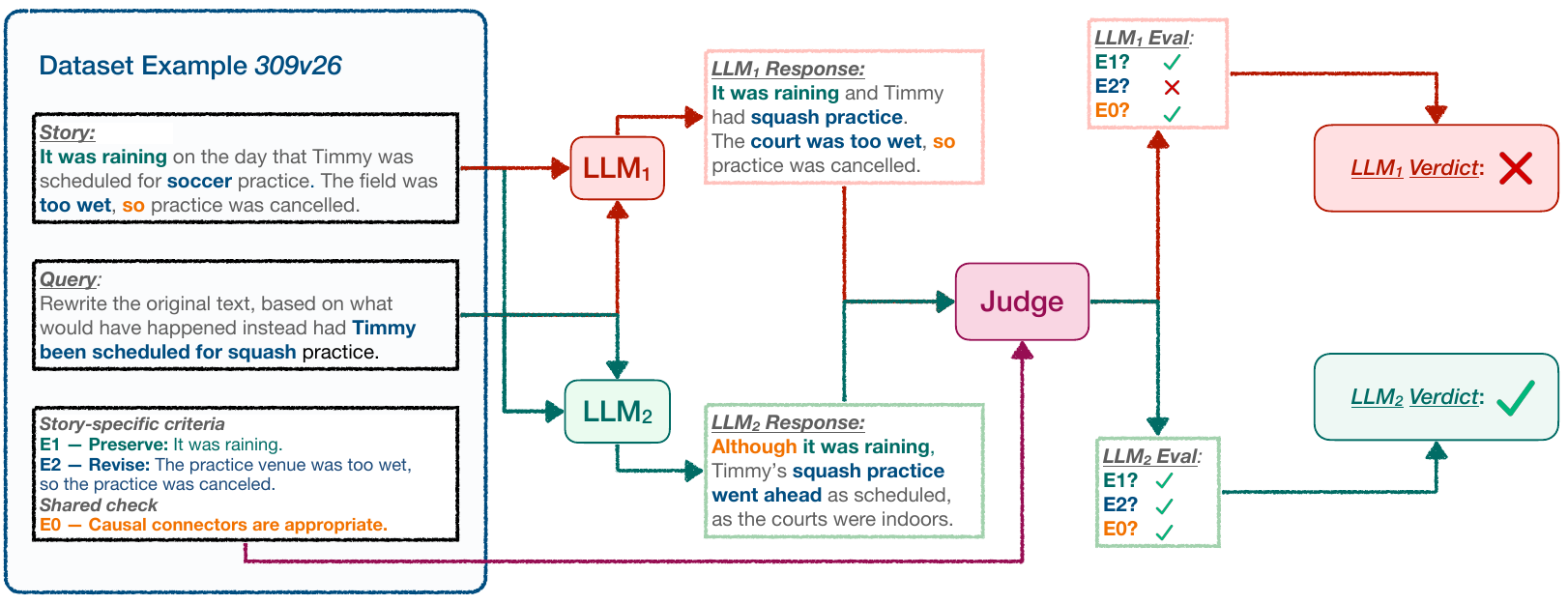}
    \vspace{-15pt}
    \caption{The two respondent LLMs illustrate a failing and a passing rewrite, evaluated independently by the same judge; a response passes only when all three checks pass.}
    \label{fig:pipeline}
\end{figure}

\section{Results} \label{sec: eval}

\paragraph{Model Suite.} We benchmark three recent closed-source models, \textit{GPT-5.4-Pro} \citep{openai2026gpt54}, \textit{Claude-Opus-4.7} \citep{anthropic2026claudeopus47}, \textit{Gemini-3.1-Pro} \citep{google2026gemini31pro}.
We include several
open-source models to get a better understanding of how much of a role scale plays in developing causal reasoning abilities. We test \textit{Llama-3.1-8B, 70B} \citep{dubey2024llama3}, \textit{Mistral-7B} \citep{jiang2023mistral}, \textit{Qwen-7B} \citep{bai2023qwen}, \textit{Phi-4} \citep{abdin2024phi4}, and 
\textit{Gemma-27B} \citep{gemmateam2024gemma2}.
Overall, the total number of models benchmarked is nine, incorporating various scales, training paradigms, and modeling techniques. For brevity, we refer to models by name only hereafter.

%% file: TR/tr_results.tex
\begin{table}[t]
\vspace{-5pt}
  \caption{Performance of models averaged across three queries.}
  \vspace{-5pt}
  \label{tab:accuracy_results}
  \centering
  \begin{adjustbox}{width=\textwidth}
  \begin{tabular}{lccccccccc}
    \toprule
    {Metric}
    & {GPT}
    & {Gemini}
    & {Claude}
    & {Llama-8B}
    & {Mistral}
    & {Qwen}
    & {Phi}
    & {Gemma}
    & {Llama-70B} \\
    \midrule
    Acc. (\%)
    & 61.34\stdv{0.45}
    & 56.68\stdv{0.44}
    & 56.14\stdv{2.50}
    & 33.77\stdv{2.50}
    & 32.90\stdv{3.36}
    & 36.83\stdv{3.41}
    & 44.73\stdv{2.02}
    & 48.26\stdv{1.45}
    & 49.64\stdv{1.42} \\
    \bottomrule
  \end{tabular}
  \end{adjustbox}
  \vspace{-10pt}
\end{table}

\begin{figure}[t!]
\centering
\includegraphics[width=1\linewidth]{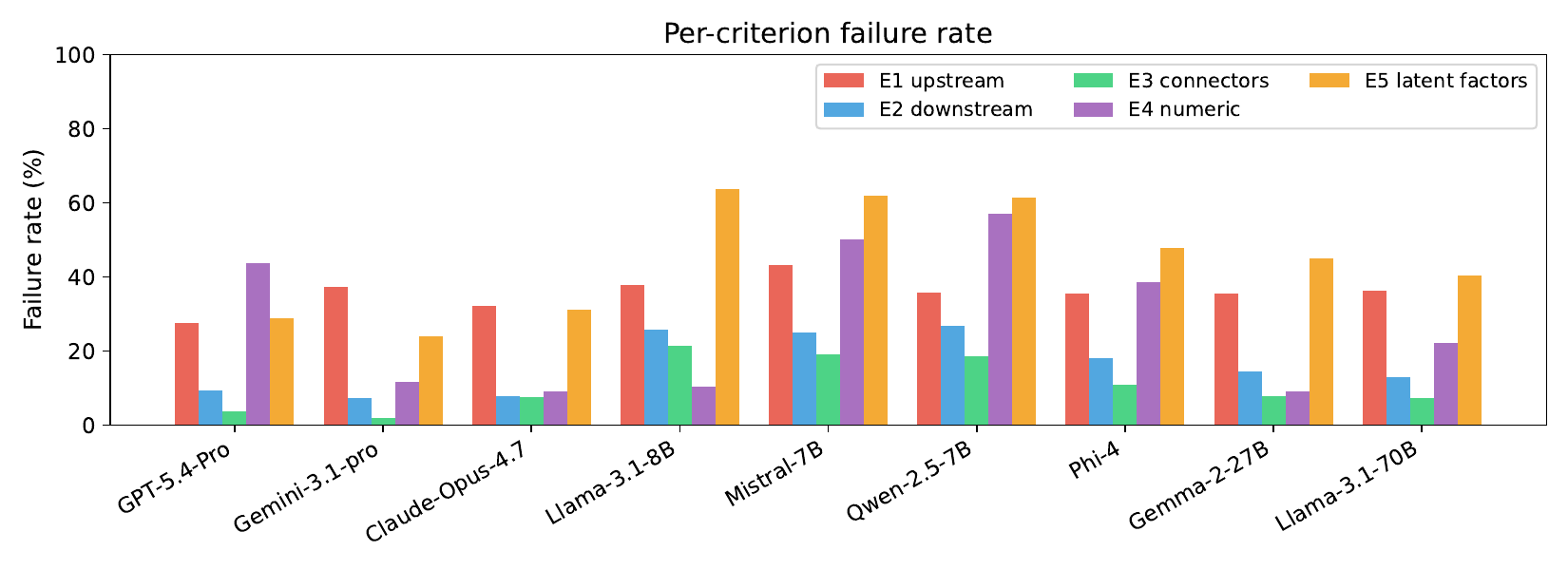}
\vspace{-10pt}
\caption{Failure rates across different evaluation criteria. E1-3 are described in Sec. \ref{subsubsec: et}. E4 and E5 are interesting sub-types where LLMs reason wrongly about the direction of a downstream edit, and fail to correctly infer latent factors implicit in the story.}
\vspace{-10pt}
\label{fig:results-error-modes}
\end{figure}

\paragraph{Overall results (\Cref{tab:accuracy_results}).}
We investigate the benchmark performance of the models averaged across three different query phrasings described in \Cref{subsec: data construction}. We observe that all models fail to show strong performance, where the highest accuracy is only 61.34$\%$, highlighting
the brittle nature of the conversational counterfactual reasoning capabilities of modern LLMs.

An intriguing observation is that the performance of LLama-70B is close to larger models (e.g., 6.5$\%$ gap with Claude), despite it having a significantly smaller number of parameters.
This implies that 
the gains on scaling the weights, the training cost and the model development cycle by several magnitudes might not 
bring improvements on counterfactual reasoning 
to the same extent. This suggests that there needs to be an explicit effort in developing systems that are natively designed to reason better, and innovations beyond scaling are required to cross the next threshold of performance.

\paragraph{Performance across different evaluation criteria (\Cref{fig:results-error-modes}).}
To analyze different types of errors in model responses, we examine the failure rates per the evaluation criteria described in \Cref{subsubsec: et}.
The general trend of models being prone to making incorrect edits to the non-descendant variables (E1 in \Cref{fig:results-error-modes}), perhaps due to mistaking correlational relationships it has observed in the training data as cause-and-effect relationships, seems consistent across the board. Comparatively, the errors due to not editing the downstream variables (E2) or using the wrong connectors (E3) are generally lower.

Among the frontier models, GPT does the worst on interventions when numerical changes are expected (E4), compared to Gemini and Claude, the latter of which does well, as does Llama. 
On investigating the errors due to not keeping latent factors invariant across the factual and counterfactual world (E5), we observe the superiority of large-scale frontier models compared to the open source models. We hypothesize this is due to larger models having a better understanding of unobserved factors in the given context, as opposed to the models with fewer parameters.

\begin{figure}[t!]
\centering
\includegraphics[width=0.95\linewidth]{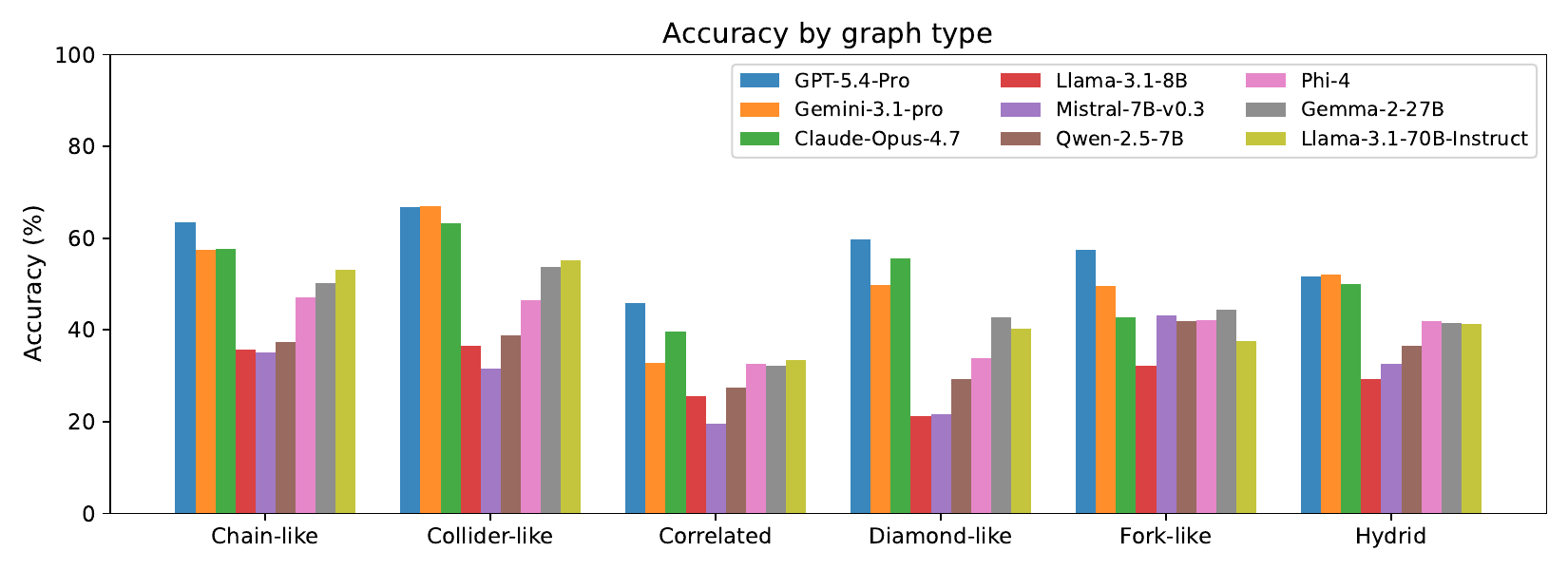}
\vspace{-5pt}
\caption{Accuracy across different graph types.}
\vspace{-12pt}
\label{fig:results-graph-types}
\end{figure}

\paragraph{Performance across different graph types (\Cref{fig:results-graph-types}).}
The differences in accuracy of different models across varied graph types show that 
the models are prone to misinterpreting correlated graphs as cause-and-effect graphs, 
illustrating the behavior of LLMs that 
primarily rely on associational knowledge. Furthermore, the performance is also lower for Diamond-like and Fork-like graphs,
indicating that models struggle
with consistently propagating effects of interventions across different downstream branches.

\begin{figure}[t!]
\centering
\includegraphics[width=0.95\linewidth]{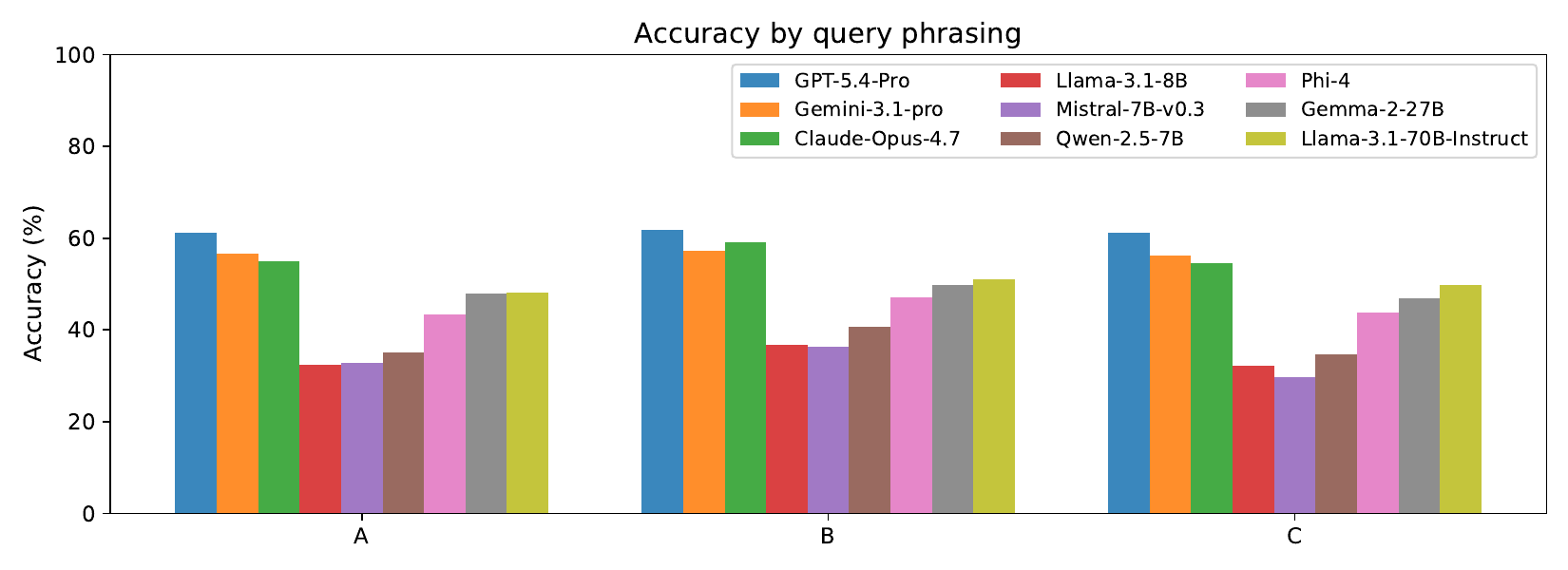}
\vspace{-10pt}
\caption{Accuracy across different query phrasings.}
\vspace{-10pt}
\label{fig:results-query}
\end{figure}

\paragraph{Performance across different query phrasings (\Cref{fig:results-query}).}
We investigate how different query phrasings 
impact the performance of models.
Looking at three paraphrased versions of the same query and comparing results per query-type illustrates that there is negligible difference in performance across different query phrasings. This corroborates that the high failure rates of models are likely due to their lack of native, conversational counterfactual reasoning capabilities.

\paragraph{Model biases.}
Among the flagship models, Claude seems to be performing the best on numerical changes, but struggles on maintaining the latent state constant for the factual and counterfactual worlds. Furthermore, while Gemini is exceptional at maintaining coherent grammatical structure, it suffers from making slightly more upstream changes. 
GPT seems to be most error prone when dealing with scenarios involving updates in numerical quantities but does better than average on refraining from changing the non-descendant variables, which suggests it might actually be less reliant on purely observational data. Different characteristics of models when sliced across different axes reveal insights that would have been hard to gain otherwise from a black box setting. This is one of the examples of how \texttt{CoBe} could be better used for understanding model behavior.

\paragraph{Prompt ablations.} We additionally show results from prompting three models with alternative prompt templates designed to provide increasingly more procedural instructions to the model. Results are in App. \ref{sec: alt_prompts}, using the same 500 random samples across prompt ablation settings. We find that providing additional conversational prompts that even more strongly suggest the counterfactual interpretation only improve performance by around 5\% on average for frontier models, and lower performance for smaller ones. We also show results from adding increasingly more explicit causal instructions to the prompt, while clarifying that this is strictly outside our task scope. We find that adding a simple clarification like \textit{"Keep prior and unaffected facts unchanged, and update the consequences of the requested change."} only improves scores by around 3\%. Finally, we show that even a fully detailed step-by-step causal recipe that elaborates this task following structural causal semantics still leaves our benchmark unsaturated, with frontier models scoring $81.3\%$ and $76.1\%$.

%% file: TR/tr_conclusion.tex
\section{Conclusion} \label{sec: conclusion}
We introduce \texttt{CoBe}, a principled and causally-grounded approach to benchmarking the ability of language models to perform conversational counterfactual text editing. We construct a diverse dataset, covering scenarios from varied domains, with different underlying graph structures and inducing novel failure modes. The dataset is structured in an easy-to-use format that should aid researchers in investigating these blind-spots and developing better systems. 

\paragraph{Limitations.} As clarified in Sec. \ref{sec: intro}, counterfactual text editing is an important but specific example of counterfactual reasoning; solving our benchmark should not be read as the sign of having achieved general counterfactual reasoning capabilities. A natural future direction is extending the benchmark to different domains, e.g., vision. A further limitation, common to benchmarks, is the possibility of author-bias in the curation of samples, which we address by ensuring the diversity in terms of domains and query phrasing, as validated by independent human quality checks.

%% file: TR/tr_appendix.tex
\newpage
\section{Background on Counterfactuals}

\subsection{Structural Causal Models and the Pearl Causal Hierarchy} \label{sec:pch}

We provide an optional primer on Pearl's structural account of causality. 
We denote random variables by capital letters, $X$, and their values by
lowercase letters, $x$. Bold letters, $\mathbf{X}$ and $\mathbf{x}$,
denote sets of variables and their corresponding values.
$P(\mathbf{x})$ is shorthand for $P(\mathbf{X}=\mathbf{x})$.

\vspace{-0.1in}
\paragraph{Structural causal models.}
We use \textit{Structural Causal Models} (SCMs) to describe the
generative process for a system of interest
\citep{pearl:2k,bareinboim:textbook}.
An SCM $\mathcal{M}$ is a tuple
$\langle \mathbf{V},\mathbf{U},\mathcal{F},P(\mathbf{u})\rangle$,
where $\mathbf{V}$ is the set of endogenous variables,
$\mathbf{U}$ is the set of exogenous variables representing
background factors outside the model, and $P(\mathbf{u})$
is their joint distribution.
$\mathcal{F}=\{f_i\}$ is a set of structural functions, each of
which causally generates the value of $V_i\in\mathbf{V}$ as
\[
    V_i \gets f_i(\mathbf{U}_i,\mathbf{Pa}_i),
\]
where $\mathbf{U}_i\subseteq\mathbf{U}$ and
$\mathbf{Pa}_i\subseteq\mathbf{V}\setminus\{V_i\}$.
We consider recursive SCMs, so these assignments can be evaluated
in an acyclic order. A realization $\mathbf{U}=\mathbf{u}$,
called a \textit{context}, determines the values of all endogenous
variables through these functions.

\vspace{-0.1in}
\paragraph{Causal graphs.}
Each SCM induces a \textit{causal graph}, or \textit{causal diagram},
$\mathcal{G}$, containing a vertex for each $V_i\in\mathbf{V}$.
A directed edge $V_j\rightarrow V_i$ is included when
$V_j\in\mathbf{Pa}_i$, and a bidirected edge
$V_i\leftrightarrow V_j$ represents dependence between their
exogenous inputs $\mathbf{U}_i$ and $\mathbf{U}_j$, including
shared unobserved causes.
Recursiveness implies that $\mathcal{G}$ has no directed cycles.
An ancestor of a variable is a vertex from which it can be reached
along a directed path; a descendant is a vertex reachable from it
along a directed path. Following our convention, these sets
include the variable itself.
The graph encodes qualitative causal dependencies; it does not,
by itself, specify the structural functions or the distribution
of background factors.

\vspace{-0.1in}
\paragraph{Interventions and potential responses.}
The operator $\operatorname{do}(\mathbf{X}=\mathbf{x})$ indexes
a submodel $\mathcal{M}_{\mathbf{x}}$ in which the functions
generating $\mathbf{X}$ are replaced by the constant assignments
$\mathbf{X}\gets\mathbf{x}$. All other structural functions and
the exogenous distribution remain unchanged.
The value of a variable $Y$ in this submodel, for context
$\mathbf{u}$, is its \textit{potential response},
denoted $Y_{\mathbf{x}}(\mathbf{u})$.
Unlike conditioning on $\mathbf{X}=\mathbf{x}$, an intervention
overrides the mechanisms that naturally determine $\mathbf{X}$.

\vspace{-0.1in}
\paragraph{Layers of the PCH.}
The \textit{Pearl Causal Hierarchy} (PCH), also known as the
\textit{ladder of causation}, distinguishes three progressively
richer regimes of reasoning
\citep{pearl:mackenzie2018,Bareinboim2022OnPH}:
\begin{enumerate}
    \item \textbf{Association ($\mathcal{L}_1$; seeing).}
    Observational quantities, such as $P(Y=y\mid X=x)$,
    describe the system under its natural mechanisms.

    \item \textbf{Intervention ($\mathcal{L}_2$; doing).}
    Interventional quantities, such as
    $P(Y=y;\operatorname{do}(X=x))=P(Y_x=y)$,
    describe the system when specified mechanisms are overridden.

    \item \textbf{Counterfactuals ($\mathcal{L}_3$; imagining).}
    Counterfactual quantities relate potential responses under
    possibly different regimes for the same unit.
    For example, $P(Y_x=y\mid X=x',Y=y')$ asks what $Y$ would
    have been had $X$ been fixed to $x$, given that $X=x'$
    and $Y=y'$ were actually observed.
\end{enumerate}
Higher layers subsume lower ones, but lower-layer distributions
do not, in general, determine higher-layer quantities without
additional causal assumptions.

\vspace{-0.1in}
\paragraph{Connection to counterfactual text editing.}
Given factual evidence $\mathbf{E}=\mathbf{e}$, counterfactual
inference proceeds through \textit{abduction}, updating the
distribution of background factors to $P(\mathbf{u}\mid\mathbf{e})$;
\textit{intervention}, constructing $\mathcal{M}_{\mathbf{x}}$;
and \textit{prediction}, evaluating the resulting potential responses.
For discrete exogenous variables and evidence of positive probability,
\[
    P(Y_{\mathbf{x}}=y\mid\mathbf{e})
    =
    \sum_{\mathbf{u}:\,Y_{\mathbf{x}}(\mathbf{u})=y}
    P(\mathbf{u}\mid\mathbf{e}).
\]
The factual and counterfactual scenarios thus share the same
background context. Non-intervened variables that are not
descendants of the intervention remain unchanged, while changes
to descendants depend on the structural mechanisms and context.
This supplies the formal basis for CoBe's intended editing task:
preserve causally unaffected facts and revise the consequences
of the requested change within the given scenario.

\subsection{Subjunctive Conditionals and Conversational Counterfactuals} \label{sec:subjunctives}

Next, we illustrate the task of conversational counterfactual text editing using the Oswald/Kennedy example from Sec. \ref{sec: intro} and Fig. \ref{fig:ctf_vs_association}. Consider a text $T=t$ describing the shooting of President Kennedy by Lee Harvey Oswald. Variables $A,B,C$ annotate events in the story.

\paragraph{Original text ($T$):} President Kennedy died ($B=0$) on Nov 22, 1963. He was shot by Lee Harvey Oswald ($A=1$), who likely acted alone ($C=0$).

\par\medskip
\begingroup
\small
\setlength{\tabcolsep}{3pt}
\renewcommand{\arraystretch}{1.2}
\noindent
\begin{tabularx}{\linewidth}{@{}
    >{\raggedright\arraybackslash}X
    >{\raggedright\arraybackslash}p{0.15\linewidth}
    >{\raggedright\arraybackslash}p{0.19\linewidth}
    >{\raggedright\arraybackslash}p{0.15\linewidth}
    >{\raggedright\arraybackslash}p{0.15\linewidth}
@{}}
\toprule
\textbf{Question} &
\textbf{Grammar} &
\textbf{Formal query} &
\textbf{Causal ladder} &
\textbf{Answer} \\
\midrule

\textbf{1.} Given that Kennedy died on Nov 22, what
\textit{would have happened} if Oswald \textit{had not}
shot him?
&
Subjunctive \textit{(would have...had})
&
\textcolor{OliveGreen}{%
$\begin{aligned}[t]
    &P(T_{A=0}\mid A=1,\\
    &\quad B=0,\,C=0)
\end{aligned}$}
&
\textcolor{OliveGreen}{Rung 3}
&
Kennedy would have been alive.
\\
\midrule

\textbf{2.} Given that Kennedy died on Nov 22, what
\textit{happened} if Oswald \textit{did not}
shoot him?
&
Indicative \textit{(did...if})
&
\textcolor{BrickRed}{%
$\begin{aligned}[t]
    &P(T\mid A=0,\\
    &\quad B=0)
\end{aligned}$}
&
\textcolor{BrickRed}{Rung 1}
&
Someone else shot Kennedy.
\\

\bottomrule
\end{tabularx}
\par
\endgroup
\medskip

This is a classic example, going back at least to Ernest Adams on the difference between subjunctive and indicative conditionals, and later discussions in the Stalnaker-Lewis tradition \citep{Adams1970,stalnaker:68,lewis:73a}. What separates the two text questions is that the first asks a \textcolor{OliveGreen}{counterfactual} question and uses a \textit{subjunctive} conditional (\textit{"what would have happened...had"}), whereas the second asks an \textcolor{BrickRed}{associational} question and uses an \textit{indicative} conditional (\textit{"what did happen...if"}). 

The use of \textbf{subjunctive or indicative language is thus the intuitive linguistic indicator of whether a natural language question is to be interpreted as a Rung 3 or Rung 1 question}, respectively \citep{vonfintel2012subjunctive,ippolito-subjunctive}. We stress that our scope is \textit{conversational} language. Humans are capable of hearing a question like \textit{"what would have happened in this scenario if I had done something different?"} and understanding that this means background factors and prior events should be kept fixed, and the intervention should only affect downstream events. They require no formal background about causal graphs or counterfactuals, nor a detailed step-by-step procedure such as "hold causal ancestors fixed" in order to answer this question. This is the native skill we evaluate in our benchmark.

\paragraph{Conversational counterfactual text edit:} Answering question 1 involves the three steps of counterfactual text editing below. This protocol is inspired by the 3-step process of \textit{Abduction-Intervention-Prediction} in Pearl's structural causal model framework described in the previous Sec. \ref{sec:pch}. 
\begin{itemize}
    \item [(a)] We situate ourselves in the given scenario ($A=1, B=0,C=0$) and hold constant the background factors that may have produced this timeline of events, such as facts about Kennedy’s public schedule and security detail;
    \item [(b)] We imagine an external intervention that miraculously stops Oswald from shooting, written symbolically as $do(A=0)$; and
    \item [(c)] We produce text $T=t'$ that reflects the likely downstream effects of this edit in the original scenario.
\end{itemize}

In this example, we are thus sampling from the distribution \textcolor{OliveGreen}{$P(T_{A=0} \mid A=1, B=0, C=0)$}: the distribution of text had we intervened with $do(A=0)$ in the original scenario having $A=1, B=0, C=0$. Notably, this edit should hold fixed any background factors and other events that are not causally downstream of the edited events, as shown in \textcolor{OliveGreen}{green} in Fig. \ref{fig:ctf_vs_association}.

\paragraph{Associational text generation:} By contrast, answering question 2 involves generating alternative text by sampling from \textcolor{BrickRed}{$P(T \mid A=0, B=0)$}. This is the distribution of text we are likely to see, conditioned on the events $A=0, B=0$, possibly updating background conditions and prior events to maintain factual correlation as shown in \textcolor{BrickRed}{red} in Fig. \ref{fig:ctf_vs_association}.

\newpage
\section{Dataset Samples}

In this section, we first provide several examples of LLM responses, along with the evaluation criteria used for the scenario, representative answer, and judgment reasoning (App.~\ref{sec:examples}).
Then, we provide a sample JSON format in our dataset (App.~\ref{subsubsec: sample json}).

\subsection{Examples}
\label{sec:examples}

The examples below illustrate preservation of unaffected facts, revision
of intervention-dependent facts, and consistency of causal connectors.
\textcolor{NavyBlue}{Blue} highlights facts to preserve;
\textcolor{Plum}{purple} marks the requested intervention;
\textcolor{OliveGreen}{green} marks appropriate revisions; and
\textcolor{BrickRed}{red} identifies the error under discussion.
Illustrative answers are not scored and need not be unique.
Author-annotated reasoning explains the intended interpretation and is
not part of the dataset.

\subsubsection{Sample Failure 1: Spatial consequence}
\label{subsubsec: fm5}
\begin{tcolorbox}[
  breakable,
  colback=CadetBlue!5, colframe=CadetBlue!65,
  colbacktitle=CadetBlue!15, coltitle=black,
  title={\textbf{Example 126v7}},
  fontupper=\small, boxrule=0.7pt, arc=2pt,
  left=6pt, right=6pt, top=5pt, bottom=5pt,
  before skip=6pt, after skip=8pt
]
\par\smallskip\noindent\textbf{Story}\quad %
Walking \textcolor{NavyBlue}{east across the Brooklyn Bridge} at 8 AM, Sato
faced into the morning sun, with the \textcolor{NavyBlue}{Manhattan skyline
at her back}.\par

\par\smallskip\noindent\textbf{Query}\quad %
\textit{Rewrite the original text, based on what would have happened
in this specific scenario instead had Sato been walking east across
the Brooklyn Bridge at \textcolor{Plum}{6 PM instead of 8 AM}.}\par

\par\smallskip\noindent\textbf{Llama3-8B response}\quad %
Walking \textcolor{NavyBlue}{east across the Brooklyn Bridge} at
\textcolor{OliveGreen}{6 PM}, Sato \textcolor{BrickRed}{faced into the fading light of the
setting sun}, with the \textcolor{NavyBlue}{Manhattan skyline at her back}.\par

\par\smallskip\noindent\textbf{Illustrative answer (not scored)}\quad %
Walking \textcolor{NavyBlue}{east across the Brooklyn Bridge} at
\textcolor{OliveGreen}{6 PM}, Sato had the \textcolor{OliveGreen}{late-afternoon sun behind
her}, with the \textcolor{NavyBlue}{Manhattan skyline at her back}.\par

\par\smallskip\noindent\textbf{Evaluation criteria}\par
\noindent\textcolor{NavyBlue}{\textbf{Should NOT change:}} Walking east across the Brooklyn Bridge, Sato had the Manhattan
skyline at her back.\par
\noindent\textcolor{Plum}{\textbf{Should change:}} It was 8 AM. The sun was in front of Sato.\par

\par\smallskip\noindent\textbf{Author-annotated reasoning (not part of the dataset)}\quad %
Changing the time leaves Sato's direction of travel and Manhattan's
relative position unchanged. A visible setting sun would be in the
western sky, behind an eastward walker rather than in front. The
illustrative answer shows a daylight realization; the criteria do
not require sunlight to remain visible at 6 PM.\par
\end{tcolorbox}

\subsubsection{Sample Failure 2: Misattributing an unchanged event}
\label{subsubsec: fm2}
\begin{tcolorbox}[
  breakable,
  colback=CadetBlue!5, colframe=CadetBlue!65,
  colbacktitle=CadetBlue!15, coltitle=black,
  title={\textbf{Example 311v21}},
  fontupper=\small, boxrule=0.7pt, arc=2pt,
  left=6pt, right=6pt, top=5pt, bottom=5pt,
  before skip=6pt, after skip=8pt
]
\par\smallskip\noindent\textbf{Story}\quad %
Vera is a highly uncharismatic politician, but a
\textcolor{NavyBlue}{major scandal ruined her opponent's reputation}, so
she won the election.\par

\par\smallskip\noindent\textbf{Query}\quad %
\textit{Rewrite the original text, based on what would have happened
in this scenario had Vera been a \textcolor{Plum}{highly charismatic}
politician and \textcolor{Plum}{lost the election} instead.}\par

\par\smallskip\noindent\textbf{Claude Opus 4.7 response}\quad %
Vera was a \textcolor{OliveGreen}{highly charismatic} politician, but a major
scandal ruined \textcolor{BrickRed}{her reputation}, so she
\textcolor{OliveGreen}{lost the election}.\par

\par\smallskip\noindent\textbf{Illustrative answer (not scored)}\quad %
Vera is a \textcolor{OliveGreen}{highly charismatic} politician, and a
\textcolor{NavyBlue}{major scandal ruined her opponent's reputation},
\textcolor{OliveGreen}{but she lost the election}.\par

\par\smallskip\noindent\textbf{Evaluation criteria}\par
\noindent\textcolor{NavyBlue}{\textbf{Should NOT change:}} A major scandal ruined her opponent's reputation.\par
\noindent\textcolor{Plum}{\textbf{Should change:}} Vera is a highly uncharismatic politician. She won the election.\par

\par\smallskip\noindent\textbf{Author-annotated reasoning (not part of the dataset)}\quad %
The query stipulates both charisma and electoral defeat; it does
not ask the model to predict the election from charisma. The
opponent's scandal is unchanged background in the intended
scenario. Reassigning the scandal to Vera changes that background
to rationalize the requested defeat.\par
\end{tcolorbox}

\subsubsection{Sample Failure 3: Retaining an inapplicable consequence}
\label{subsubsec: fm3}
\begin{tcolorbox}[
  breakable,
  colback=CadetBlue!5, colframe=CadetBlue!65,
  colbacktitle=CadetBlue!15, coltitle=black,
  title={\textbf{Example 309v26}},
  fontupper=\small, boxrule=0.7pt, arc=2pt,
  left=6pt, right=6pt, top=5pt, bottom=5pt,
  before skip=6pt, after skip=8pt
]
\par\smallskip\noindent\textbf{Story}\quad %
It was \textcolor{NavyBlue}{raining} and Timmy had soccer practice. The field
was too wet, so practice was canceled.\par

\par\smallskip\noindent\textbf{Query}\quad %
\textit{Rewrite the original text, based on what would have happened
in this scenario had Timmy had \textcolor{Plum}{squash practice} instead.}\par

\par\smallskip\noindent\textbf{GPT 5.4 Pro response}\quad %
It was \textcolor{NavyBlue}{raining} and Timmy had \textcolor{OliveGreen}{squash}
practice. The court was \textcolor{BrickRed}{too wet, so practice was canceled}.\par

\par\smallskip\noindent\textbf{Illustrative answer (not scored)}\quad %
It was \textcolor{NavyBlue}{raining}, and the outdoor soccer field was still
too wet. Timmy had \textcolor{OliveGreen}{squash practice} instead, and it
\textcolor{OliveGreen}{went ahead on an indoor court}.\par

\par\smallskip\noindent\textbf{Evaluation criteria}\par
\noindent\textcolor{NavyBlue}{\textbf{Should NOT change:}} It was raining.\par
\noindent\textcolor{Plum}{\textbf{Should change:}} The practice venue was too wet, so the practice was canceled.\par

\par\smallskip\noindent\textbf{Author-annotated reasoning (not part of the dataset)}\quad %
Here, ``practice venue'' refers to the venue used for the activity
in the rewritten scenario. Under the ordinary indoor-squash
interpretation, rain making the outdoor soccer field too wet is
no longer a reason to cancel that practice. The response transfers
the wet-venue explanation to the squash court. The outdoor field
need not become dry, and the criterion does not rule out every
other possible reason for cancellation.\par
\end{tcolorbox}

\subsubsection{Sample Failure 4: Unsupported causal connector}
\label{subsubsec: fm6}
\begin{tcolorbox}[
  breakable,
  colback=CadetBlue!5, colframe=CadetBlue!65,
  colbacktitle=CadetBlue!15, coltitle=black,
  title={\textbf{Example 115v3}},
  fontupper=\small, boxrule=0.7pt, arc=2pt,
  left=6pt, right=6pt, top=5pt, bottom=5pt,
  before skip=6pt, after skip=8pt
]
\par\smallskip\noindent\textbf{Story}\quad %
The structural engineer observed that the primary support beam
had a \textcolor{NavyBlue}{three-inch fissure across its center}. Due to the
\textcolor{NavyBlue}{high risk of immediate collapse}, the city council
ordered the building to be evacuated immediately.\par

\par\smallskip\noindent\textbf{Query}\quad %
\textit{Rewrite the original text, based on what would have happened
instead had the city council decided to \textcolor{Plum}{keep the
building open for business, rather than order an evacuation}.}\par

\par\smallskip\noindent\textbf{Illustrative model response (scored)}\quad %
The structural engineer observed that the primary support beam
had a \textcolor{NavyBlue}{three-inch fissure across its center}.
\textcolor{BrickRed}{Due to} the \textcolor{NavyBlue}{high risk of immediate collapse},
the city council decided to \textcolor{OliveGreen}{keep the building open for
business, rather than order an evacuation}.\par

\par\smallskip\noindent\textbf{Illustrative answer (not scored)}\quad %
The structural engineer observed that the primary support beam
had a \textcolor{NavyBlue}{three-inch fissure across its center}.
\textcolor{OliveGreen}{Despite} the \textcolor{NavyBlue}{high risk of immediate collapse},
the city council decided to \textcolor{OliveGreen}{keep the building open for
business, rather than order an evacuation}.\par

\par\smallskip\noindent\textbf{Evaluation criteria}\par
\noindent\textcolor{NavyBlue}{\textbf{Should NOT change:}} The support beam had a three-inch fissure. There was a high risk
of collapse.\par
\noindent\textcolor{Plum}{\textbf{Should change:}} The city council ordered the building to be evacuated immediately.\par

\par\smallskip\noindent\textbf{Author-annotated reasoning (not part of the dataset)}\quad %
The intervention changes the council's decision, not the observed
damage or the stated risk. The response incorporates that decision
but retains ``due to,'' presenting the danger as the reason to
keep the building open. That connection does not follow from the
given scenario. ``Despite'' expresses the intended contrast;
neutral wording without a causal connector would also be valid.
No collapse or injury is required.\par
\end{tcolorbox}

\subsubsection{Sample Failure 5: Overlooking an unchanged purpose}
\label{subsubsec: fm1}
\begin{tcolorbox}[
  breakable,
  colback=CadetBlue!5, colframe=CadetBlue!65,
  colbacktitle=CadetBlue!15, coltitle=black,
  title={\textbf{Example 403v1}},
  fontupper=\small, boxrule=0.7pt, arc=2pt,
  left=6pt, right=6pt, top=5pt, bottom=5pt,
  before skip=6pt, after skip=8pt
]
\par\smallskip\noindent\textbf{Story}\quad %
I am going to the shop to \textcolor{NavyBlue}{repair my umbrella}. It is
raining heavily outside. I \textcolor{NavyBlue}{took my umbrella} when I
went out.\par

\par\smallskip\noindent\textbf{Query}\quad %
\textit{Rewrite the original text, based on what would have happened
in this scenario had it \textcolor{Plum}{not been raining at all} instead.}\par

\par\smallskip\noindent\textbf{GPT 5.4 Pro response}\quad %
I am going to the shop to \textcolor{NavyBlue}{repair my umbrella}. It is
\textcolor{OliveGreen}{not raining outside at all}, so I
\textcolor{BrickRed}{did not take my umbrella} when going out.\par

\par\smallskip\noindent\textbf{Illustrative answer (not scored)}\quad %
I am going to the shop to \textcolor{NavyBlue}{repair my umbrella}. It is
\textcolor{OliveGreen}{not raining outside at all}. I still
\textcolor{NavyBlue}{took my umbrella} when I went out, since I needed to
get it repaired.\par

\par\smallskip\noindent\textbf{Evaluation criteria}\par
\noindent\textcolor{NavyBlue}{\textbf{Should NOT change:}} I am going to the shop to repair my umbrella. I took my umbrella
when I went out.\par
\noindent\textcolor{Plum}{\textbf{Should change:}} It is raining heavily outside.\par

\par\smallskip\noindent\textbf{Author-annotated reasoning (not part of the dataset)}\quad %
The repair errand remains unchanged and supplies a reason to
bring the umbrella even without rain. The response changes the
weather correctly but treats rain as the only reason for carrying
the umbrella, overlooking the purpose explicitly stated in the story.\par
\end{tcolorbox}

\subsubsection{Sample Failure 6: Altering a historical fact}
\label{subsubsec: fm11}
\begin{tcolorbox}[
  breakable,
  colback=CadetBlue!5, colframe=CadetBlue!65,
  colbacktitle=CadetBlue!15, coltitle=black,
  title={\textbf{Example 419v1}},
  fontupper=\small, boxrule=0.7pt, arc=2pt,
  left=6pt, right=6pt, top=5pt, bottom=5pt,
  before skip=6pt, after skip=8pt
]
\par\smallskip\noindent\textbf{Story}\quad %
Hari's favorite color is red. He discovers a
\textcolor{NavyBlue}{childhood photo of his girlfriend wearing a red dress}.
He \textcolor{NavyBlue}{had never seen this photo before}.\par

\par\smallskip\noindent\textbf{Query}\quad %
\textit{Rewrite the original text, based on what would have happened
instead had \textcolor{Plum}{Hari's favorite color been blue instead
of red}.}\par

\par\smallskip\noindent\textbf{Claude Opus 4.7 response}\quad %
Hari's favorite color is \textcolor{OliveGreen}{blue}. He discovers a childhood
photo of his girlfriend wearing a \textcolor{BrickRed}{blue dress}. He
\textcolor{NavyBlue}{had never seen this photo before}.\par

\par\smallskip\noindent\textbf{Illustrative answer (not scored)}\quad %
Hari's favorite color is \textcolor{OliveGreen}{blue}. He discovers a
\textcolor{NavyBlue}{childhood photo of his girlfriend wearing a red dress}.
He \textcolor{NavyBlue}{had never seen this photo before}.\par

\par\smallskip\noindent\textbf{Evaluation criteria}\par
\noindent\textcolor{NavyBlue}{\textbf{Should NOT change:}} Hari discovers a previously unseen childhood photo of his
girlfriend. She is wearing a red dress in the photo.\par
\noindent\textcolor{Plum}{\textbf{Should change:}} Hari's favorite color is red.\par

\par\smallskip\noindent\textbf{Author-annotated reasoning (not part of the dataset)}\quad %
The requested change concerns Hari's preference, not the historical
contents of the photograph. Changing the dress to blue makes it
match his new preference but alters an unaffected fact. The
criteria make no demand about his emotional reaction to the photo.\par
\end{tcolorbox}

\subsubsection{Sample Failure 7: Violating a quantitative constraint}
\label{subsubsec: fm77}
\begin{tcolorbox}[
  breakable,
  colback=CadetBlue!5, colframe=CadetBlue!65,
  colbacktitle=CadetBlue!15, coltitle=black,
  title={\textbf{Example 317v3}},
  fontupper=\small, boxrule=0.7pt, arc=2pt,
  left=6pt, right=6pt, top=5pt, bottom=5pt,
  before skip=6pt, after skip=8pt
]
\par\smallskip\noindent\textbf{Story}\quad %
The water level in the bathtub is rising rapidly because the
\textcolor{NavyBlue}{main faucet is pouring water in at a rate of 5 gallons
per minute}.\par

\par\smallskip\noindent\textbf{Query}\quad %
\textit{Rewrite the original text, based on what would have happened
instead had \textcolor{Plum}{the drain been opened, leaking water at
8 gallons per minute}.}\par

\par\smallskip\noindent\textbf{Illustrative model response (scored)}\quad %
The water level in the bathtub is \textcolor{BrickRed}{rising rapidly} because
the \textcolor{NavyBlue}{main faucet is pouring water in at a rate of
5 gallons per minute}. The \textcolor{OliveGreen}{drain was opened, leaking
water at 8 gallons per minute}.\par

\par\smallskip\noindent\textbf{Illustrative answer (not scored)}\quad %
The \textcolor{NavyBlue}{faucet continues to pour water into the bathtub at
5 gallons per minute}, while the \textcolor{OliveGreen}{opened drain removes
8 gallons per minute}. The water level \textcolor{OliveGreen}{falls} as the tub
loses a net \textcolor{OliveGreen}{3 gallons per minute}.\par

\par\smallskip\noindent\textbf{Evaluation criteria}\par
\noindent\textcolor{NavyBlue}{\textbf{Should NOT change:}} The main faucet is pouring water in at a rate of 5 gallons per minute.\par
\noindent\textcolor{Plum}{\textbf{Should change:}} The water level in the bathtub is rising rapidly.\par

\par\smallskip\noindent\textbf{Author-annotated reasoning (not part of the dataset)}\quad %
The faucet's inflow remains fixed, and the intervention adds an
outflow exceeding it. While water remains and the stated rates
hold, the net volume change is $5-8=-3$ gallons per minute, so the
water level falls. The response includes the requested drain
opening but incorrectly retains the original rising level.\par
\end{tcolorbox}

\subsubsection{Sample Success: Marked correct despite multiple rewrites}
\label{subsubsec: ss1}
\begin{tcolorbox}[
  breakable,
  colback=CadetBlue!5, colframe=CadetBlue!65,
  colbacktitle=CadetBlue!15, coltitle=black,
  title={\textbf{Example 217v10}},
  fontupper=\small, boxrule=0.7pt, arc=2pt,
  left=6pt, right=6pt, top=5pt, bottom=5pt,
  before skip=6pt, after skip=8pt
]
\par\smallskip\noindent\textbf{Story}\quad %
Using a paper bookmark to keep her place, Lisa reads a thick novel
\textcolor{NavyBlue}{at the terminal}. Her connecting flight has been
\textcolor{NavyBlue}{delayed by three hours}.\par

\par\smallskip\noindent\textbf{Query}\quad %
\textit{Rewrite the original text, based on what would have happened
in this specific scenario instead had Lisa \textcolor{Plum}{watched
a movie on her laptop}.}\par

\par\smallskip\noindent\textbf{Gemini 3.1 Pro response}\quad %
Three alternative rewrites are shown below; introductory text and
option headings are omitted.\par

\begin{enumerate}[leftmargin=1.5em,itemsep=3pt,topsep=3pt]
  \item Using a pair of headphones to block out the noise, Lisa
  \textcolor{OliveGreen}{watches a movie on her laptop} \textcolor{NavyBlue}{at the
  terminal}. Her connecting flight has been \textcolor{NavyBlue}{delayed
  by three hours}.
  \item Hitting the spacebar to pause her screen whenever
  announcements chime, Lisa \textcolor{OliveGreen}{watches a movie on her
  laptop} \textcolor{NavyBlue}{at the terminal}. Her connecting flight has
  been \textcolor{NavyBlue}{delayed by three hours}.
  \item Sitting close to a wall outlet to keep her battery charged,
  Lisa \textcolor{OliveGreen}{watches a movie on her laptop}
  \textcolor{NavyBlue}{at the terminal}. Her connecting flight has been
  \textcolor{NavyBlue}{delayed by three hours}.
\end{enumerate}

\par\smallskip\noindent\textbf{Illustrative answer (not scored)}\quad %
Lisa \textcolor{OliveGreen}{watches a movie on her laptop}
\textcolor{NavyBlue}{at the terminal}. Her connecting flight has been
\textcolor{NavyBlue}{delayed by three hours}.\par

\par\smallskip\noindent\textbf{Evaluation criteria}\par
\noindent\textcolor{NavyBlue}{\textbf{Should NOT change:}} Lisa is at the terminal. Her connecting flight has been delayed
by three hours.\par
\noindent\textcolor{Plum}{\textbf{Should change:}} Lisa is reading a thick novel and using a paper bookmark to keep
her place in that reading activity.\par

\par\smallskip\noindent\textbf{Author-annotated reasoning (not part of the dataset)}\quad %
Each alternative replaces the foregrounded reading activity with
watching a movie while preserving the location and flight delay.
Headphones, pausing, and charging are optional elaborations, not
required consequences. The change does not require Lisa to
discard the bookmark; it may remain in the closed book.\par
\end{tcolorbox}

\subsection{Sample JSON}
\label{subsubsec: sample json}

\begin{tcblisting}{
  listing only,
  breakable,
  colback=CadetBlue!5,
  colframe=CadetBlue!30,
  colbacktitle=CadetBlue!15,
  coltitle=black,
  title={\textbf{Sample JSON entry: 317v3}},
  boxrule=0.5pt,
  arc=2pt,
  left=6pt, right=6pt, top=5pt, bottom=5pt,
  listing options={
    basicstyle=\ttfamily\footnotesize,
    columns=fullflexible,
    keepspaces=true,
    showstringspaces=false,
    breaklines=true,
    breakatwhitespace=true,
    breakautoindent=false,
    breakindent=1.5em,
    postbreak={},
    aboveskip=0pt,
    belowskip=0pt
  }
}
{
  "Core Set ID": 317,
  
  "Variation ID": 3,
  
  "Domain": "Everyday",
  
  "Story": "The water level in the bathtub is rising rapidly because the main faucet is pouring water in at a rate of 5 gallons per minute.",
  
  "Query": [
    "Rewrite the original text, based on what would have happened instead had the drain been opened, leaking water at 8 gallons per minute.",
    "Based on the preceding text, rewrite the scenario to reflect what would have happened if the drain had been opened, leaking water at 8 gallons per minute.",
    "Rewrite the above passage to illustrate what would have occurred had the drain been opened, leaking water at 8 gallons per minute."
  ],
  
 "Illustrative answer (not scored)": "The faucet continues to pour water into the bathtub at 5 gallons per minute, while the opened drain removes 8 gallons per minute. The water level falls as the tub loses a net 3 gallons per minute.",
  
  "Evaluation criteria": [
    "The main faucet is pouring water in at a rate of 5 gallons per minute.",
    "The water level in the bathtub is rising rapidly."
  ]
}
\end{tcblisting}

\noindent
The entry above illustrates the dataset schema. The two strings in
\texttt{Evaluation criteria} list, respectively, facts to preserve and
original facts to revise. The \texttt{Illustrative answer (not scored)}
provides one example of a valid rewrite; multiple valid answers are possible.

Here, the faucet's inflow remains at 5 gallons per minute. The requested
intervention introduces an outflow of 8 gallons per minute, so the water
level falls rather than rises.

\newpage
\section{Qualitative Failure Modes}
\label{subsubsec: fm}

The following qualitative patterns elaborate the evaluation checks
in Sec.~\ref{subsubsec: et}. They describe errors in the rewritten
text rather than establishing the internal reasoning process that
produced them. The patterns can overlap and do not introduce
additional scoring criteria. The linked examples illustrate each
pattern.

\begin{enumerate}[leftmargin=15pt,itemsep=4pt]
    \item \textbf{Overlooking an unchanged reason for an action.}
    A rewrite removes an action after changing one associated
    condition, even though the story supplies another reason for
    the action that remains in place. In
    \hyperref[subsubsec: fm1]{403v1}, the umbrella is still needed
    for the repair errand when the rain is removed.

    \item \textbf{Changing background facts to rationalize the intervention.}
    A rewrite alters an unaffected event to make a stipulated
    outcome appear more expected. In
    \hyperref[subsubsec: fm2]{311v21}, the query specifies both
    greater charisma and electoral defeat, but the response
    reassigns the opponent's scandal to Vera.

    \item \textbf{Retaining a consequence whose stated reason no longer applies.}
    A rewrite incorporates the requested intervention but carries
    over an explanation tied to the original situation. In
    \hyperref[subsubsec: fm3]{309v26}, the response transfers the
    wet-venue cancellation from outdoor soccer to the
    counterfactual squash session, under the intended indoor
    interpretation.

    \item \textbf{Violating physical or quantitative constraints.}
    A rewrite retains an outcome incompatible with the conditions
    in the counterfactual scenario. In
    \hyperref[subsubsec: fm77]{317v3}, an inflow of 5 gallons per
    minute and an outflow of 8 gallons per minute imply a falling
    water level while the stated flows hold, not a rising one.

    \item \textbf{Updating spatial consequences inconsistently.}
    A rewrite changes an event but fails to keep its consequences
    consistent with the preserved spatial setting. In
    \hyperref[subsubsec: fm5]{126v7}, an eastward walker cannot
    face a setting sun in the western sky. The criterion does not
    require the sun to remain visible at 6 PM.

    \item \textbf{Using an unsupported causal connector.}
    A rewrite changes the requested event but retains wording that
    assigns it an unsupported cause. In
    \hyperref[subsubsec: fm6]{115v3}, ``due to the high risk of
    immediate collapse'' is carried over as the explanation for
    keeping the building open. The issue is the asserted causal
    relationship, not the absence of a particular replacement
    word such as ``despite.''

    \item \textbf{Changing historical facts to match a preference.}
    A rewrite makes an unaffected past event agree with a newly
    stipulated preference. In
    \hyperref[subsubsec: fm11]{419v1}, changing Hari's favorite
    color does not change the red dress recorded in his
    girlfriend's childhood photograph.
\end{enumerate}

\section{Evaluation Details} \label{sec: eval appendix}

In this section, we first describe the infrastructural details on the evaluation in App.~\ref{app:evaluation-details-infrastructure}, and then provide the prompts used for assessing each evaluation criterion in App.~\ref{subsec: llm-as-a-judge prompts}. We also provide the details of human agreement in App.~\ref{appendix:human}, and then discuss additional results in App.~\ref{app:additional-results}.

\subsection{Infrastructure}
\label{app:evaluation-details-infrastructure}
For the open-source models the responses were generated using 1-4 NVIDIA H100 GPUs. For the frontier models, we generated the responses using API keys.
Wherever \textit{thinking-mode} or \textit{reasoning-effort} or an equivalent parameter is available we activate it and set it to \textit{medium} or the \textit{default} applicable.

\subsection{LLM-Judge prompts}
\label{subsec: llm-as-a-judge prompts}

Our LLM-Judge evaluation pipeline checks the supplied criteria rather
than matching responses to a single reference answer. It is designed
to accommodate differences in wording and presentation, including
responses with additional text or multiple valid rewrites (e.g.,
\hyperref[subsubsec: ss1]{217v10}). Q0 examines connector usage in the
rewrite, while Q1 and Q2 assess the rewrite against the first and
second evaluation-criterion strings, respectively. Agreement with
human judgments is reported in App.~\ref{appendix:human}.

The prompts provided to the LLM-Judge are shown below.
\texttt{\$response\_text} denotes the model's rewrite, and
\texttt{\$eq1} and \texttt{\$eq2} denote the first and second
strings in \texttt{Evaluation criteria}.

\begingroup
\tcbset{
  listing only,
  breakable,
  width=\linewidth,
  colback=CadetBlue!5,
  colframe=CadetBlue!50,
  colbacktitle=CadetBlue!15,
  coltitle=black,
  boxrule=0.6pt,
  arc=2pt,
  left=6pt, right=6pt, top=5pt, bottom=5pt,
  before skip=8pt, after skip=8pt,
  listing options={
    language={},
    basicstyle=\ttfamily\footnotesize,
    columns=fullflexible,
    keepspaces=true,
    showstringspaces=false,
    numbers=none,
    breaklines=true,
    breakatwhitespace=false,
    breakautoindent=false,
    breakindent=1em,
    postbreak={},
    aboveskip=0pt,
    belowskip=0pt,
    literate={→}{{$\rightarrow$}}1 {é}{{\'{e}}}1
  }
}

\begin{tcblisting}{title={\textbf{Q0: Wrong connectors}}}
Criterion: "Every causal connector in the rewrite accurately reflects the causal relationship described."

Mentally list each causal connector (so, thus, therefore, but, however, consequently, as a result, etc.).

Reply F if *any* causal connector misrepresents the relationship between the connected words. This includes either: the causal relationship is in the wrong or reversed direction, or the connected words are not causally related.

Reply T if *every* causal connectors signals the correct causal direction. If the causal relationship is reasonably plausible, reply T. If there are no causal connectors, reply T.

Short examples (your entire output is exactly one letter):

Example 1: rewrite says "Due to the storm, the school did not declare a snow day. Because they were in class during the day, the kids were not able to go outside and go sledding while the sun was up." → F
Example 2: rewrite says "Because it was a home derby at the Bernabéu, Real Madrid wore their purple away kit." → F
Example 3: rewrite says "The weather was hot, so Alex grabbed a hot tea" → F
Example 4: rewrite says "The weather was hot, so Alex grabbed an iced coffee" → T
Example 5: rewrite says "Because the weather was sunny, Alice brought her umbrella" → F
Example 6: rewrite says "Despite the sunny weather, Alice brought her umbrella" → T

Rewrite:
$response_text

Your entire reply must be exactly one character: T or F.
\end{tcblisting}

\begin{tcblisting}{title={\textbf{Q1: Preservation of invariant facts}}}
Criterion: "Facts that should be UNCHANGED"

Check whether every claim in the criterion is preserved in the rewrite with the same meaning.

Reply T if *all* criterion claims are still present and factually unchanged, even if they are paraphrased or expressed differently.

Reply F if *any* criterion claim is omitted, contradicted, or materially changed.

Short examples (your entire output is exactly one letter):

Example 1: criterion says pipe leaking and floor wet; rewrite keeps leak but floor is dry → F
Example 2: criterion says "There was a defect at the factory"; rewrite says "Without the factory defect ..." → T
Example 3: criterion says "Jake scored the goal"; rewrite says "Amy scored the goal" → F
Example 4: criterion says "David is driving a car"; rewrite says "A car ran out of gas because david drove the car for 10 hours" → T
Example 5: criterion says "Defenders successfully blocked the attack."; rewrite says "Defenders detected the intrusion too late. Despite the defenders' attempt to block the attack, they allowed the opponent to score the goal" → F

Criterion (facts that should be UNCHANGED):
"$eq1"

Rewrite:
$response_text

Your entire reply must be exactly one character: T or F.
\end{tcblisting}

\begin{tcblisting}{title={\textbf{Q2: Required changes}}}
Criterion: "Facts that should be CHANGED"

Check whether every claim in the criterion has been changed so that it no longer holds in the rewrite with the same substantive meaning.

Reply T if *every* criterion claim is contradicted, removed, or otherwise changed so that the original claim no longer remains true.

Reply F if *any* criterion claim is still preserved and continues to hold in the rewrite, even if it is paraphrased or embedded in a longer statement.

Short examples (your entire output is exactly one letter):

Example 1: criterion says "brakes failed. car hit wall."; rewrite says "brakes engaged and car was safe" → T
Example 2: criterion says "Jake scored the goal"; rewrite says "Amy scored the goal" → T
Example 3: criterion says "David drove the car"; rewrite says "The car ran out of gas because David drove it for 10 hours" → F
Example 4: criterion says "The government ordered the street to be evacuated"; rewrite says "The government chooses to keep the street open instead of evacuation" → T
Example 5: criterion says "Alice pours water on the table. She cleaned the water."; rewrite says "Alice was surprised that she poured milk on the table. She cleaned the milk" → T
Example 6: criterion says "Defenders successfully blocked the attack."; rewrite says "Defenders detected the intrusion too late, allowing the opponent to score the goal" → T
Example 7: criterion says "Drinking a cold water, Jake ate a spicy kimchi"; rewrite says "Drinking a cold water, Jake ate a mild tofu" → F
Example 8: criterion says "test was hard for Alice"; rewrite says "test would not have been hard for Alice if she tried hard" → T

Criterion (facts that should be CHANGED):
"$eq2"

Rewrite:
$response_text

Your entire reply must be exactly one character: T or F.
\end{tcblisting}

\endgroup

\newpage
\subsection{Dataset Construction and Quality Assurance} \label{appendix:QA}
\subsubsection{Sample curation and augmentation}

The dataset construction ran in four stages:

\renewcommand{\arraystretch}{1.3}
\begin{longtable}{@{}P{0.17\textwidth} P{0.17\textwidth} P{0.29\textwidth} P{0.31\textwidth}@{}}
\caption{Dataset construction pipeline.}
\label{tab:dataset-construction} \\

\toprule
\textbf{Stages (1-4)} &
\textbf{Output} &
\textbf{Steps taken} &
\textbf{Quality assurance and filtration} \\
\midrule
\endfirsthead

\midrule
\multicolumn{4}{r}{\small\textit{Continued on next page}} \\
\endfoot

\bottomrule
\endlastfoot

\textbf{Counterfactual example collection}
&
\textbf{425 candidate counterfactual examples}
&
A cohort of student researchers constructed examples following the dataset schema in Sec.~3. Each example contained an original scenario, a counterfactual edit request, and explicit evaluation criteria for LLM answers.
&
All contributors had completed a graduate causal inference course. Each example was required to specify a coherent counterfactual intervention, the resulting edit, and criteria identifying which facts should remain fixed and which consequences should change.
\\[0.8em]

\textbf{Core-set selection}
&
\textbf{106 diverse core examples crafted}
&
The authors manually reviewed the 425 candidates and used them as a basis to craft a core set spanning diverse domains, event interactions, and causal structures.
&
We rejected candidates with ill-defined counterfactuals, ambiguous or unsound causal relationships, incoherent scenarios, or evaluation criteria that did not clearly distinguish valid from invalid edits.
\\[0.8em]

\textbf{Augmentation of core set}
&
\textbf{25 candidate variations per core example}
&
Each core example was augmented using LLMs to produce variations across diverse themes and domains while preserving the original causal structure and logical relationships among events.
&
Generation was repeated until 25 satisfactory variations were obtained for each core example. We rejected and regenerated variations that were incoherent or contained unsound evaluation criteria.
\\[0.8em]

\textbf{Final verification and filtering}
&
\textbf{2,650 examples; 2,000 final examples retained after quality checks}
&
The authors manually reviewed every set of 25 augmented examples and its associated evaluation criteria.
&
Every augmented set of 25 variations was checked for story coherence and for the correctness and clarity of its evaluation criteria. Only examples passing this review were included in the final dataset. External expert feedback led to additional examples being removed.
\\

\end{longtable}
\renewcommand{\arraystretch}{1}

\paragraph{Augmentation and development checks.}
Augmentation used Gemini-3.1-Pro and GPT-5.4-Pro through
separate accounts without conversational memory. The authors
reviewed generated variations and their evaluation criteria
before inclusion. During development, we also spot-checked
selected examples on language models. These checks did not
determine inclusion: examples were retained even when model
responses were correct, or were initially incorrect but later became correct with model updates. 
Selection prioritized scenario quality, diversity of domains and causal
structures, and clarity of the evaluation criteria.

\paragraph{Author review and filtering.}
At stages 2 and 4 above,
each item was reviewed by at least two authors. Review covered
scenario coherence, clarity of the requested intervention,
and the correctness of the facts designated for preservation
or revision. Authors used color-coded markings to indicate
approval and disagreement, and discussed disagreements before
a final decision. We only retained examples where both reviewing authors came to a consensus that the
example and its criteria were acceptable. We do not report a quantitative
inter-author agreement statistic for this stage.

Grounds for rejection included ill-defined counterfactuals,
ambiguous or unsound causal relationships, incoherent
scenarios, and evaluation criteria that did not clearly
distinguish acceptable from unacceptable edits. Initial
author filtering was followed by external expert review
and additional exclusions informed by that review.
Together, these checks removed 650 incorrect or weak examples of the 2,650 augmented
examples, leaving 2,000 final scenarios.

\subsubsection{External Expert Quality Audit}

We recruited three experts in causal inference who had not
participated in dataset construction. After initial author
filtering, we randomly sampled 100 examples from the remaining
candidate pool. All three experts independently reviewed
the same 100 examples, assessing coherence, grammatical
quality, and validity of the counterfactual request on a
five-point scale. They also assessed the diversity of the
reviewed collection and provided written feedback. The experts were not shown the preservation and revision
criteria. This review assessed scenario and query quality,
complementing the authors' separate verification of those
criteria. The questions they answered were,

\begin{itemize}
    \item \textbf{Coherence}: Does the story present a clear and logical flow of events? Is its overall structure coherent?
    \item \textbf{Grammar}: Is the text grammatically natural, intelligible, and easy to read?
    \item \textbf{Counterfactual validity}: Does the query express a valid counterfactual intervention on the events described in the story?
    \item \textbf{Dataset diversity}: Taken together, do the reviewed examples represent a diverse range of scenarios, domains, and causal structures?
    \item \textbf{Additional feedback (text)}: Are there any other issues or observations that should be brought to our attention?
\end{itemize}

Interestingly, expert feedback identified a systematic
inconsistency and led to further exclusions of examples
exhibiting that issue. Namely, the experts noticed that some examples contained a causal relationship that was contrived. We noticed this was due to a systemic issue from a batch of augmented samples. These exclusions are included in the total of 650 reported above.

Across all 100 audited examples, mean scores were \textbf{4.85/5}
for coherence, \textbf{4.94/5} for grammatical quality, and \textbf{4.71/5}
for counterfactual-query validity. The reviewed collection
received a mean diversity rating of \textbf{5/5}. These scores include
examples subsequently excluded and therefore describe the
audited candidate sample before the final exclusions.

Reviewers further noted that the dataset covers \textit{"social life, physics, biology, chemistry, sports and other domains"} and contains \textit{"a good mix of random daily life scenarios and science ones and a few calculation ones too with a good variety of graph types and intervention locations."}

\newpage
\subsection{Human Agreement} \label{appendix:human}
\subsubsection{Setup}
To assess agreement between our LLM judge and independent human
judgments, we randomly sampled $250$ model responses, each paired
with its original story and counterfactual edit query, from the
outputs of one model in our evaluation suite. Three expert human
annotators independently assessed each response using the general
rubric below. Annotators were shown the story, query, and model
response, but not the story-specific \texttt{Evaluation criteria}.
They were provided with two worked examples illustrating the
intended counterfactual interpretation. The majority human judgment
served as the reference label for comparison with the LLM judge.

\begingroup
\tcbset{
  listing only,
  breakable,
  width=\linewidth,
  colback=CadetBlue!5,
  colframe=CadetBlue!50,
  colbacktitle=CadetBlue!15,
  coltitle=black,
  boxrule=0.6pt,
  arc=2pt,
  left=6pt, right=6pt, top=5pt, bottom=5pt,
  before skip=8pt, after skip=8pt,
  listing options={
    language={},
    basicstyle=\ttfamily\footnotesize,
    columns=fullflexible,
    keepspaces=true,
    showstringspaces=false,
    numbers=none,
    breaklines=true,
    breakatwhitespace=false,
    breakautoindent=false,
    breakindent=1em,
    postbreak={},
    aboveskip=0pt,
    belowskip=0pt,
    literate={’}{{'}}1 {“}{{"}}1 {”}{{"}}1
  }
}

\begin{tcblisting}{title={\textbf{Human-rating instructions}}}
Hello!

Thanks for agreeing to be one of the human raters for helping us understand the counterfactual reasoning capabilities of modern large language models. The goal of this activity is to get human opinions on outputs produced by querying LLMs with certain interesting counterfactual queries.

You would be provided with 250 short conversations with an LLM, and you are to judge whether the LLM output is correct. In each conversation you will be provided with the following:

Original Context (T = t): a paragraph describing a scenario with some events and outcomes.

Query (T_x = t’ | T = t, X = x’): A natural language question that asks how the original story (T) would have changed if one of the original events/conditions (X = x’) was instead fixed by external intervention to be different (X = x).

LLM Output (T = t’’): the response from a model to the original context and query.

For each conversation, your task is to judge whether the LLM Output (T=t’’) is a valid answer to the Query (T\_x = t’ | T = t, X = x’). In other words, if the original story is altered by external intervention, would the LLM Output be a valid counterfactual scenario?

Note: while the Query has a fixed technical interpretation, there could be multiple valid answers to this Query. You are to evaluate whether the LLM Output is reasonable given the original context.
\end{tcblisting}

\begin{tcblisting}{title={\textbf{Human-rating rubric and examples}}}
Please mark "Yes", if you think the LLM Output is
(a) factually consistent with the Original Context; (b) incorporates the external intervention mentioned in the Query;
(c) appropriately reflects downstream changes following from the intervention, without affecting causal non-descendants; and
(d) does not change any causal implications from the Original Context, such as by using wrong connectors (e.g. “due to X” instead of “despite X”).

Please mark "No", if you think the LLM Output is (a) factually inconsistent with the Original Context; or (b) does not appropriately incorporate the external intervention from the Query, or reflect this edit in causally downstream events; or (c) changes events that are not causal descendants of the intervention; or (d) changes causal implications from the Original Context, such as by using wrong connectors (e.g. “due to X” instead of “despite X”); or (e) hallucinates or produces incoherent output.

Please use the "Other" option sparingly and add comments and justification, whenever a clear Yes/No answer is not provided.

EXAMPLE 1

Original Context: Due to the storm, the school declared a snow day. Because they were not in class during the day, the kids were able to go outside and go sledding while the sun was up.

Query: Rewrite the original text, based on what would have happened instead had they not been granted the snow day.

LLM Output: Despite the storm, the school did not declare a snow day. Because they were in class during the day, the kids were unable to go outside and go sledding while the sun was up.

Human Evaluation: YES - the LLM Output is correct. It incorporates the intervention and downstream changes, does not change non-descendants, and correctly edits “due to” to “despite”.

EXAMPLE 2

Original Context: Due to the clear skies, the hawk was able to see and catch the mouse from a distance. As a result, the wolf was unable to find something to eat that night.

Query: Rewrite the original text, based on what would have happened instead had the wolf been able to find food.

LLM Output: Due to the cloudy skies, the hawk was unable to see and catch the mouse from a distance. As a result, the wolf was able to find something to eat that night.

Human Evaluation: NO - the LLM Output is wrong. It changes prior events when incorporating the intervention.
\end{tcblisting}

\endgroup

\subsubsection{LLM Judge Alignment and Statistical Significance}
We then processed the same $250$ sampled instances through our LLM-as-a-judge pipeline using a few-shot prompt.

The LLM achieved an approximate \textit{Accuracy} of $92.1$\%, \textit{Precision} of $82.5\%$, \textit{Recall} of $91.7$\% and \textit{F1 Score} of $86.81\%$ (where failure is the positive class) against the human majority. More importantly, to account for random chance in a binary classification task, we calculated Cohen’s $\kappa$ between the LLM and the human majority, resulting in $\kappa = 0.812$. This demonstrates that the LLM achieves a level of inter-rater reliability with the consensus that is \textit{almost-perfect}. Given the sample size ($N=250$), this agreement provides sufficient statistical confidence to scale the evaluator across the broader benchmark. Further, the high agreement is strong evidence that our evaluation criteria that we use in the dataset are themselves sound.

\begin{figure}[t!]
\centering
\includegraphics[width=\linewidth]{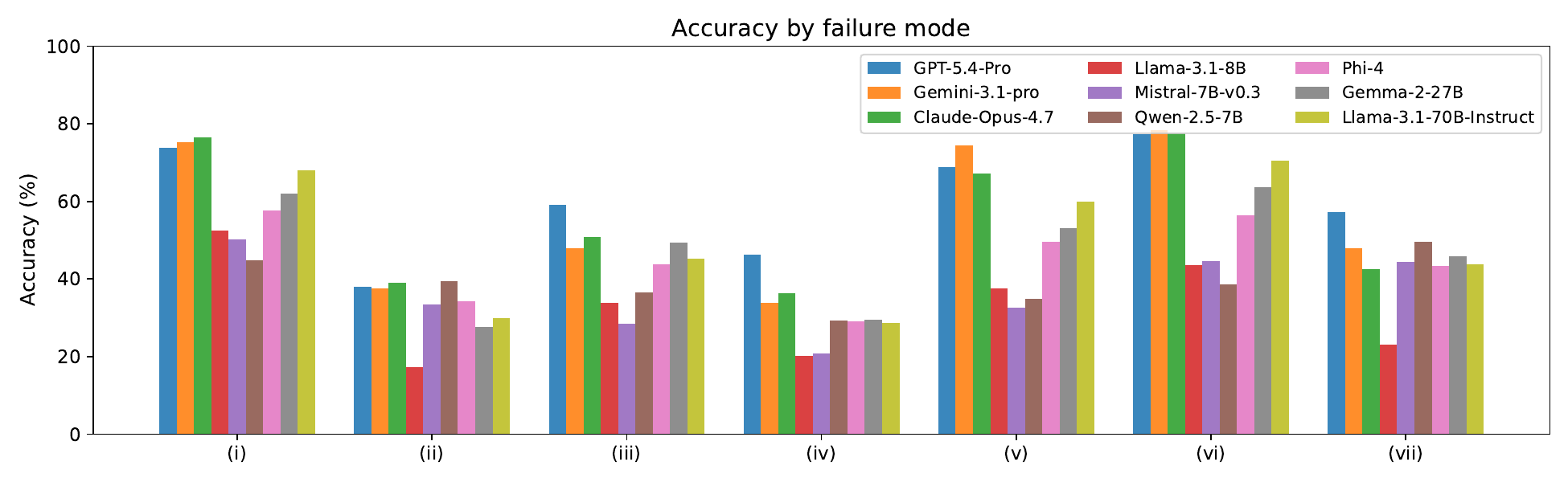}
\caption{Accuracy across different failure modes.}
\label{fig:results-failure-modes}
\end{figure}

\subsection{Additional results}\label{app:additional-results}

\paragraph{Performance across different failure modes (\Cref{fig:results-failure-modes}).} Investigating the performance of various models across the fine-grained failure modes defined in Sec.~\ref{subsubsec: fm} reveals interesting patterns. LLMs struggle in maintaining invariant the scientific laws established either explicitly or implicitly in the scenario while making counterfactual edits.

As expected, the LLMs tend to mistake correlational relationships for cause-and-effect relationships and showcase a significant failure rate in scenarios where there is a correlational relationship with the potential to be misinterpreted.

Likewise, there are several instances of failures in scenarios where the effect of the intervention has to be propagated to the downstream variables but the LLMs fail to do so appropriately, highlighting that they still struggle to effectively update the scenario to map to a valid counterfactual world resulting from the intervention. This observation also corroborates the higher than average failure rate for the Diamond- and Fork-like graphs as noted in Fig.~\ref{fig:results-graph-types}.

\section{Prompt Ablation Experiments} \label{sec: alt_prompts}

We evaluate four tiers of prompting: the original benchmark prompts
and three additional ablations that provide progressively more
explicit guidance about the intended counterfactual edit.

\begin{itemize}[leftmargin=15pt]
    \item \textbf{Tier 1 (A--C): Original conversational prompts.}
    The three query templates in Sec.~\ref{sec: data} specify a
    hypothetical change and ask the model to rewrite the story,
    leaving the intended editing procedure implicit.

    \item \textbf{Tier 2 (D--G): More explicit conversational prompts.}
    The templates in App.~\ref{subsec: explicit_prompts} more strongly
    cue an intervention in the given scenario, while leaving the
    preservation and updating rules implicit. They remain within
    our conversational task setting.

    \item \textbf{Tier 3 (H1): Explicit preservation and updating instructions.}
    The template in App.~\ref{subsec: prompt_engineering} explicitly
    instructs the model to preserve prior and unaffected facts and
    update the consequences of the requested change. This supplies
    an editing rule that the model must infer in the primary task.

    \item \textbf{Tier 4 (H2): A detailed causal recipe.}
    The template in App.~\ref{subsec: causal_recipe} provides a
    step-by-step procedure using causal terminology, including
    graph construction, intervention, and updating causal descendants.
\end{itemize}

Our scope distinction concerns whether the prompt supplies the
editing rules, rather than simply whether it uses conversational
language. Tiers 1 and 2 leave these rules implicit; Tiers 3 and 4
provide explicit guidance and are therefore outside the primary
task setting defined in Sec.~\ref{sec: intro}. We nevertheless
report them as instruction-assisted controls to examine how
performance changes when some of the inference burden is removed.

We evaluated all four tiers on the same $500$ randomly sampled
examples for GPT-5.4-Pro, Gemini-3.1-Pro, and Qwen-2.5-7B.
Table~\ref{tab:prompt-engineering} reports their pass rates, with
scores averaged over the corresponding templates for Tiers 1 and 2.

\begin{table}[t]
\centering
\caption{Pass rates across four prompt tiers on the same $500$
randomly sampled examples. A--C and D--G are conversational
prompts; H1 supplies explicit preservation and updating instructions,
and H2 supplies a detailed causal recipe.}
\label{tab:prompt-engineering}
\label{tab:prompt-comparison}
\small
\setlength{\tabcolsep}{6pt}
\renewcommand{\arraystretch}{1.15}
\begin{adjustbox}{max width=\linewidth}
\begin{tabular}{lcccc}
\toprule
Model & Tier 1 & Tier 2 & Tier 3 & Tier 4 \\
& A--C & D--G & H1 & H2 \\
\midrule
GPT-5.4-Pro    & 66.2\% & 71.7\% & 73.5\% & 81.3\% \\
Gemini-3.1-Pro & 58.7\% & 64.4\% & 67.1\% & 76.1\% \\
Qwen-2.5-7B    & 36.2\% & 33.3\% & 25.0\% & 32.0\% \\
\bottomrule
\end{tabular}
\end{adjustbox}
\end{table}

\subsection{More Explicit Conversational Prompts (Tier 2)}
\label{subsec: explicit_prompts}

We provide four additional templates that strengthen the intended
interventionist interpretation without specifying which facts to
preserve or giving a causal procedure. These phrasings are designed
to discourage alternative readings such as \textit{backtracking
counterfactuals}, in which background conditions may be revised
to accommodate the hypothesized event
\citep{vonKugelgen2023backtracking}. In particular, references to
an external intervention or to going back in time and altering
an event emphasize a change within the given scenario. These
templates retain the requirement that the model infer how to
carry out the edit from the conversational request.

\paragraph{Additional prompts.}
\begin{itemize}[leftmargin=20pt]
    \item[\textit{(D)}] \textit{Rewrite the above text to illustrate
    what would have happened in this specific scenario, had
    \textless{}intervention happened\textgreater{}.}

    \item[\textit{(E)}] \textit{Rewrite the above text to incorporate
    an imagined external intervention in this scenario, where
    instead of \textless{}original scenario happened\textgreater{},
    \textless{}intervened scenario happened\textgreater{}.}

    \item[\textit{(F)}] \textit{Rewrite the story described above,
    had we gone back in time and altered
    \textless{}original scenario happened\textgreater{} to
    \textless{}intervened scenario happened\textgreater{}.}

    \item[\textit{(G)}] \textit{Rewrite the above text based on how
    events would have unfolded, but where instead of
    \textless{}original scenario happened\textgreater{},
    \textless{}intervened scenario happened\textgreater{}.}
\end{itemize}

\paragraph{Results.}
Averaging over D--G, GPT improves from $66.2\%$ to $71.7\%$
($+5.5$ percentage points) and Gemini from $58.7\%$ to $64.4\%$
($+5.7$ points). Qwen instead declines from $36.2\%$ to $33.3\%$
($-2.9$ points). Thus, stronger conversational cues provide
moderate gains for the two frontier models but do not consistently
improve performance across the models tested. Substantial failure
rates persist under these alternative phrasings.

\subsection{Explicit Preservation and Updating Instructions (Tier 3)}
\label{subsec: prompt_engineering}

Template H1 explicitly states the preservation and updating rules
that are left implicit in Tiers 1 and 2. Although the instruction
uses everyday language, it supplies part of the intended editing
policy rather than requiring the model to infer it from the
counterfactual request. It is therefore outside our primary task
setting, but provides a useful control for the effect of explicit
task guidance.

\begin{tcolorbox}[
    breakable,
    colback=CadetBlue!5,
    colframe=CadetBlue!50,
    colbacktitle=CadetBlue!15,
    coltitle=black,
    boxrule=0.6pt,
    arc=2pt,
    fontupper=\small,
    title=\textbf{Template H1: Explicit editing instructions}
]
\textit{Original story: ...}

\medskip
\textit{Rewrite prompt: Rewrite the original text based on the
counterfactual scenario in which \texttt{<intervention>}.}

\medskip
\textit{Keep prior and unaffected facts unchanged, and update the
consequences of the requested change.}

\medskip
\textit{Your final answer should consist only of the rewritten sentence.}
\end{tcolorbox}

\paragraph{Results.}
Relative to Tier 1, H1 raises the reported pass rates of GPT and
Gemini by $7.3$ and $8.4$ percentage points, respectively. The
additional gains over Tier 2 are smaller: $1.8$ and $2.7$ points.
For Qwen, performance instead falls to $25.0\%$, below both
conversational tiers. Explicitly stating the editing rules
therefore has a model-dependent effect and does not uniformly
resolve the observed failures.

\subsection{A Detailed Causal Recipe (Tier 4)}
\label{subsec: causal_recipe}

Template H2 provides a detailed procedure grounded in structural
causal models. Unlike the conversational templates, it explicitly
specifies how to interpret the requested change and how to update
the story. This evaluates performance with a supplied causal
procedure; the model must still infer the scenario's causal
structure and apply that procedure correctly.

\begin{tcolorbox}[
    breakable,
    colback=CadetBlue!5,
    colframe=CadetBlue!50,
    colbacktitle=CadetBlue!15,
    coltitle=black,
    boxrule=0.6pt,
    arc=2pt,
    fontupper=\small,
    title=\textbf{Template H2: Detailed causal recipe}
]
\textit{Original story: ...}

\medskip
\textit{Perform the following steps:}
\begin{enumerate}[
    leftmargin=18pt,
    label=(\arabic*),
    itemsep=3pt,
    topsep=4pt
]
    \item \textit{Identify the key variables in the original story
    above, following the semantics of Pearl's structural causal models;}

    \item \textit{Provide the causal graph connecting these variables
    and add any unobserved confounders that might be present;}

    \item \textit{The rewrite prompt below specifies a change in the
    story; treat this change as an exogenous intervention to be
    interpreted in the sense of the do(V=v) operation in causal
    semantics; update the variables in the graph under this
    intervention with the causal effects cascading only to causal
    descendants, leaving graphical non-descendant variables unchanged;}

    \item \textit{Provide the full text of the rewritten counterfactual
    story under this intervention.}
\end{enumerate}

\medskip
\textit{Rewrite prompt: Rewrite the original text based on the
counterfactual scenario in which \texttt{<intervention>}.}

\medskip
\textit{Your final answer should consist only of the rewritten
counterfactual story.}
\end{tcolorbox}

\paragraph{Results.}
With the detailed causal recipe, GPT reaches $81.3\%$ and Gemini
reaches $76.1\%$, increases of $7.8$ and $9.0$ percentage points
over H1. Qwen improves from $25.0\%$ to $32.0\%$, but remains
below its performance under both conversational tiers.
The strongest result, GPT's $81.3\%$ pass rate, still leaves
$18.7\%$ of evaluated responses failing the benchmark checks.
Performance therefore remains below saturation in this evaluation,
even when a detailed causal procedure is provided.

Together, these experiments distinguish robustness to conversational
phrasing from performance under explicit task guidance. Supplying
the causal procedure can substantially improve performance for
some models, but those gains are obtained under a different level
of assistance from the primary conversational task.